\pdfoutput=1
\documentclass{acm_proc_article-sp}
\usepackage{amsmath}
\usepackage{amssymb}
\usepackage{fancyvrb}
\usepackage{multicol}
\usepackage{graphicx}
\usepackage{url}
\usepackage{tabularx}
\usepackage{caption}
\usepackage{color}
\usepackage{subcaption}
\usepackage{epstopdf}
\usepackage{boxedminipage}
\usepackage{enumitem}
\newtheorem{thm}{Theorem}

\newtheorem{assumption}{Condition}

\begin{document}
\title{Collaborative Filtering with Information-Rich and Information-Sparse Entities}
\author{Kai Zhu, Rui Wu, Lei Ying and R. Srikant}
\numberofauthors{2} 
\author{
\alignauthor
Kai Zhu\\
 \affaddr{School of Electrical, Computer and Energy Engineering}\\
       \affaddr{Arizona State University}\\
       \affaddr{Tempe, AZ 85287}\\
       \email{kzhu1708@asu.edu}
\alignauthor
Rui Wu\\
       \affaddr{Department of Electrical and Computer Engineering}\\
       \affaddr{University of Illinois at Urbana-Champaign}\\
       \affaddr{Urbana, IL 61801}\\
       \email{ruiwu1@illinois.edu}
\and
\alignauthor
Lei Ying\\
 \affaddr{School of Electrical, Computer and Energy Engineering}\\
       \affaddr{Arizona State University}\\
       \affaddr{Tempe, AZ 85287}\\
       \email{lei.ying.2@asu.edu}
\alignauthor
R. Srikant\\
       \affaddr{Department of Electrical and Computer Engineering}\\
       \affaddr{University of Illinois at Urbana-Champaign}\\
       \affaddr{Urbana, IL 61801}\\
       \email{rsrikant@illinois.edu}
}

\date{}
\maketitle
\begin{abstract}
In this paper, we consider a popular model for collaborative filtering in recommender systems where some users of a website rate some items, such as movies, and the goal is to recover the ratings of some or all of the unrated items of each user. In particular, we consider both the clustering model, where only users (or items) are clustered, and the co-clustering model, where both users and items are clustered, and further, we assume that some users rate many items (information-rich users) and some users rate only a few items (information-sparse users). When users (or items) are clustered, our algorithm can recover the rating matrix with $\omega(MK \log M)$ noisy entries while $MK$ entries are necessary, where $K$ is the number of clusters and $M$ is the number of items. In the case of co-clustering, we prove that $K^2$  entries are necessary for recovering the rating matrix, and our algorithm achieves this lower bound within a logarithmic factor when $K$ is sufficiently large. We compare our algorithms with a well-known algorithms called alternating minimization (AM), and a similarity score-based algorithm known as the popularity-among-friends (PAF) algorithm by applying all three to the MovieLens and Netflix data sets. Our co-clustering algorithm and AM have similar overall error rates when recovering the rating matrix, both of which are lower than the error rate under PAF. But more importantly, the error rate of our co-clustering algorithm is significantly lower than AM and PAF in the scenarios of interest in recommender systems: when recommending a few items to each user or when recommending items to users who only rated a few items (these users are the majority of the total user population).  The performance difference increases even more when noise is added to the datasets.

\end{abstract} 

\section{Introduction}

Many websites today use recommender systems to recommend items of interests to their users. Well known examples include Amazon, Netflix and MovieLens, where each user is suggested items that he or she may like, using partial knowledge about all the users' likes and dislikes. In this paper, we focus on the so-called Netflix or MovieLens model in which there are a large number of users and a large number of movies (called items in this paper), and each user rates a subset of the items that they have watched. These ratings are typically from a discrete set; for example, each item could be given a rating of $1$ through $5.$ If one views the user ratings as a matrix, with users as the rows and the items as the columns, then the resulting rating matrix is typically very sparse. The reason is that most users rate only a few items. The goal of a recommender system in such a model is to recommend items that a user may like, using the sparse set of available ratings. While the real goal is to just recommend a few items that each user would like, mathematically the problem is often posed as a \emph{matrix completion} problem: fill in all the unknown entries of the matrix. The use of partial knowledge of about other users' preferences to make a prediction about a given user's preference is referred to as collaboration, and the process of making predictions is called filtering; therefore, recommender systems which use multiple users' behaviors to predict each user's behavior is said to use \emph{collaborative filtering}.

With no assumptions, the matrix completion problem is practically impossible to solve. In reality, it is widely believed that the unknown matrix of all the ratings has a structure that can be exploited to solve the matrix completion problem. The two most common assumptions about the rating matrix are the following:

\vskip -1em
\paragraph{Low-rank assumption} The assumption here is that the rating matrix has a small rank. Suppose that there are $U$ users and $M$ items, then the true rating matrix ${\bf B}$ is an $U\times M$ matrix. The low rank assumption means that the rank of the matrix $\bf B$ is assumed to be $K<<\min\{U,M\}.$ This assumption is typically justified by recalling a well-known result in linear algebra which states that every $U\times M$ matrix of rank $K$ can be written in the form ${\bf A}{\bf D}^T,$ where ${\bf A}$ is an $U\times K$ matrix and ${\bf D}$ is an $M\times K$ matrix. Thus, the $(i,j)^{\rm th}$ entry of the rating matrix ${\bf B},$ can be written as $b_{ij}={\bf a}_i{\bf d}_j^T,$ where ${\bf a}_i$ is the $i^{\rm th}$ row of ${\bf A}$ and ${\bf d}_j$ is the $j^{\rm th}$ row of ${\bf D}.$ Since ${\bf a}_i$ and ${\bf d}_j$ are vectors of length $K,$ the low-rank assumption can be viewed as follows: each user and item can be characterized by $K$ features each, and the rating of a item by a user is simply the inner product of these features. The low-rank assumption is popular because it is viewed as a mathematical abstraction of the real-life situation in which a typical user looks for only a features of a movie (such as the lead actors, director, language, genre, etc.) before he/she decides to watch it.

\vskip -1em
\paragraph{Cluster assumption} The assumption here is that users and items are grouped into clusters, such that users in the same cluster will provide similar ratings to items in the same cluster. If there are $K$ user and item clusters, and each user in a cluster provides the same rating to each item in a cluster, then the rating matrix can be summarized by a $K\times K$ block matrix. Thus, the cluster assumption would then be a stronger assumption than low-rank assumption since the rating matrix would only have $K$ independent rows (and columns) and is thus also a rank-$K$ matrix. However, if the stronger assumption leads to lower complexity algorithms with better predictive power, then such an assumption is well-justified. We emphasize that the true ratings are unknown, and so neither of the two assumptions can be actually verified in a real-life data set. The only way to justify an assumption is by studying the performance of algorithms resulting from the assumption, by using some of the known ratings as training data to predict the remaining known ratings.

\subsection{Prior Work}

We now briefly review some of the algorithms that have resulted from the above assumptions. One way to exploit the low rank assumption is to find a matrix whose rank is the smallest among all matrices which agree with the observed ratings at the known entries of the matrix. However, the resulting rank minimization problem is not a convex problem, and a popular heuristic is to replace the rank minimization objective with a nuclear norm minimization objective. In \cite{CandesRecht,CandesTao,Recht,Gross,KeshavanMontanariOh}, it was shown that, under some conditions, both the nuclear norm minimization problem and the rank minimization problem yield the same result. Additionally, the nuclear norm minimization problem is a convex problem, and therefore, polynomial-time algorithms exist to solve such problems \cite{KeshavanMontanariOh,Candesstudent}. In reality, each step of such algorithms requires one to perform complicated matrix operations (such as the Singular Value Decomposition or SVD) and many such steps are required for convergence, and hence such algorithms are too slow in practice. An alternative algorithm which runs much faster and is also quite popular in practice is the so-called alternating minimization (AM) algorithm: initially, an SVD of the known ratings portion of the rating matrix is performed to estimate the rank, and the ${\bf A}$ and ${\bf D}$ matrices mentioned earlier. Then, ${\bf A}$ is assumed to be fixed and a ``better" ${\bf D}$ is obtained by minimizing the squared error between the entries of ${\bf A}{\bf D}^T$ and the known entries of the rating matrix. Then, the resulting ${\bf D}$ is taken to be fixed, and a better ${\bf A}$ is selected using the same criterion. This process is repeated until some convergence criterion is satisfied. Thus, the AM algorithm simply alternates between improving the estimate of ${\bf A}$ and the estimate of ${\bf D},$ and hence its name. This algorithm has been known for a while, but no performance guarantees were known till recently. Recently, it has been theoretically established in \cite{SujaySTOC} that a variant of the AM algorithm obtains the true ratings under the same so-called \emph{incoherence} assumption as in the case of nuclear norm minimization, and only requires slightly more entries to guarantee exact recovery of the rating matrix. Thus, for practical purposes, one can view the AM algorithm as the best algorithm known to date for the matrix completion problem under the low-rank assumption.

If the cluster assumption is made, then one can apply popular clustering methods such as $K$-means or spectral clustering to first cluster the users. Then, to predict the rating of a particular user, one can use a majority vote among users in his/her cluster. Such an approach has been used in \cite{TomozeiMassoulie}.  In \cite{Dabeer}, for each user, a subset of other users is selected to represent the user's cluster. The subset is chosen by computing a similarity score for each pair of users, and identifying the top $k$ users who are most similar to the given user. Then the user's rating for a particular user is computed by a majority vote among his/her similarity subset. Note that neither of these algorithms uses the item clusters explicitly, and therefore, their performance could be far from optimal. Recalling our earlier comment that clustered models can be thought of as a special case of low-rank models, one can apply the low-rank matrix completion algorithms to the cluster models as well. Such an approach was taken in \cite{ourNIPSsubmission}, where it was shown that the mathematics behind the low-rank matrix completion problem can be considerably simplified under the cluster assumption. The paper, however, does not exploit the presence of information-rich users (or items) in the system.

\subsection{Our Results}

The focus of this paper is on the matrix completion problem under the cluster assumption. We consider both the clustering and co-clustering models, i.e., our results are applicable to the case when only users (or items) are assumed to be clustered and to the case when both items and users are assumed to be clustered. While the former case is considered for completeness and comparison to prior work, we believe that the latter assumption is even better based on experiments with real datasets which we report later in the paper. \emph{The significant departure in our model compared to prior work is the assumption that there are information-rich and information-sparse users and items.} In particular, we assume that there exist some users who rate a significantly larger number of items than others, and we call the former \emph{information-rich} and the latter \emph{information-sparse}. We borrowed this terminology from \cite{BanerjeeMassoulie} who use it in the context of their privacy and anonymity model. The presence of information-rich and information-sparse users in data sets has been well known. For example, in the MovieLens dataset, 38 out of 6,040 users rated more than 1,000 movies (the dataset has 3,952 movies in total), but more than 73\% of the users rated fewer than 200 movies. To the best of our knowledge, the presence of information-rich entities has not been exploited previously for matrix completion.

Our main contributions are as follows:
\begin{enumerate}[leftmargin=*]

\item We present a clustered rating matrix model in which each cluster has information-rich and information-sparse entities, where an entity here could mean a user or an item. We note that an information-rich user may only rate a small fraction of the total number of items, but this fraction is distinctly greater than that of an information-sparse user. A similar comment applies to information-rich items as well.

\item We devise a similarity based scheme as in \cite{Dabeer} to exploit the presence of information-rich users to dramatically improve the performance of the algorithm in \cite{Dabeer}. Two remarks are in order here: first, our algorithm uses a normalization, not found in \cite{Dabeer}, which allows us compare users (and items) with widely different numbers of ratings. The second remark is that, by exploiting the presence of information-rich users, the performance of our algorithm achieves an easily provable lower-bound with a logarithmic factor when we only assume that users are clustered. In the case of co-clustering, the lower bound is achieved within a logarithmic factor if the number of clusters is sufficiently large.

\item As mentioned earlier, in practice, it is hard to verify what assumptions a true rating matrix will satisfy since the matrix is unknown except for a sparse subset. In particular, even if the cluster assumption holds, it is not clear whether each subset will contain an information-rich user and an information-sparse user. Therefore, using the theory developed here, we present a heuristic algorithm which exploits the cluster assumption, but does not require each cluster to have an information-rich user. This algorithm, which is our recommended algorithm, combines the theory presented in this paper with the algorithm in \cite{Dabeer}.

\item We compare our algorithm with the AM algorithm, and the similarity score-based algorithm in \cite{Dabeer} known as the PAF (popularity-among-friends) algorithm by applying all three to both the MovieLens and Netflix datasets. As can be seen from the results presented later in the paper, our algorithm performs significantly better.

\item The proposed algorithm has low computational complexity. Consider the case where the number of items $M$ is equal to the number of users. Further let $\alpha$ denote the fraction of entries in the ratings matrix that are known, and $C$ denote the size of each user/item cluster. Then the similarity score for each user pair requires $\alpha^2 M$ computations. Since there are $M(M-1)$ similarity scores to be computed for the users, and another $M(M-1)$ scores for items, the similarity score computation for co-clustering requires $O(\alpha^2 M^3)$ computations. Further, since sorting $M$ items requires $O(M\log M)$ computations, identifying user and item clusters require a total of $O(\alpha M^3+M^2\log M)$ computations. Finally, majority voting in each $C\times C$ block requires $\alpha C^2$ operations. So the overall complexity is $O(\alpha^2 M^3+M^2\log M+\alpha M^2 C^2).$  Here one of the three terms $O(\cdot)$ will dominate, depending on the assumptions regarding the order of $\alpha$ and $C.$   Since $\alpha$ is typically small (please see the performance evaluation section for actual numbers), the computational complexity of our algorithm is quite small.

\item The proposed algorithm can be easily implemented in a distributed and parallel fashion. The pair-wise similarity scores can be computed in parallel, and the majority voting to determining the ratings can also be done simultaneously.

\item Another notable advantage of our algorithm (which is also the case for the algorithm in \cite{Dabeer}) compared to the algorithms in \cite{Candesstudent,KeshavanMontanariOh,TomozeiMassoulie} is that our algorithm can handle incremental additions to the rating matrix easily. For example, if a few new rows and/or columns are added to the matrix, then it is easy for our algorithm to compute additional similarity scores and obtain a similarity set of each user/item. This would be more computationally burdensome in the case of the any other algorithm surveyed in this paper, since one has to rerun the algorithm on the entire rating matrix.
\end{enumerate}

\section{Model}
\label{sec: model}
In this section, we present the basic model. A summary of notations can be found in Appendix \ref{sec: notation}. We consider a recommendation system consisting of $U$ users and $M$ items, and $U$ and $M$ are at the same order ($U=\Theta(M)$). Let $\bf B$ denote the $U\times M$ preference matrix, where $b_{um}$ is the preference level of user $u$ to item $m.$  We assume that there are $G$ different levels of preference, so $b_{u m}\in \{1, \cdots, G\}.$

The goal of the recommendation system is to recover the preference matrix $\bf B$ from a sparse and noisy version of the matrix, named the rating matrix $\bf R.$ We assume the users, or the items, or both are clustered. When the users are assumed to be clustered, they form $K$ user-clusters, each with size $U/K.$ We assume $K=O(M/\log M).$ In other words, the cluster size is at least of the order of $\log M.$ Without loss of generality, we assume users are indexed in such a way that user $u$ is in user-cluster $\lceil \frac{u}{U/K} \rceil=\lceil \frac{uK}{U} \rceil.$ We further assume users in the same cluster have the same preference to every item, i.e.,
$b_{u m}= b_{v m},\forall m,\hbox{if} \left\lceil\frac{uK}{U}\right\rceil=\left\lceil\frac{vK}{U}\right\rceil.$
Furthermore, we say that the preference matrix $\bf B$ is {\em fractionally separable for users} if the following condition holds.
\begin{assumption} ({\bf Fractional Separability Condition for Users}) There exists a constant $0<\mu<1$ such that
\begin{eqnarray*}
&\sum_{m=1}^M {\bf 1}_{b_{um}=b_{vm}} \leq \mu M &\quad \hbox{ if }\quad \left\lceil\frac{uK}{U}\right\rceil\not=\left\lceil\frac{vK}{U}\right\rceil\\
&\sum_{m=1}^M {\bf 1}_{b_{um}=b_{vm}} = M &\quad \hbox{ if }\quad \left\lceil\frac{uK}{U}\right\rceil =\left\lceil\frac{vK}{U}\right\rceil
\end{eqnarray*}
In other words, for any pair of users $u$ and $v$ who are not in the same cluster, they have the same preference on at most $\mu$ fraction of the items; and for any pair of users in the same cluster, they have the same preference on all items.  \hfill{$\square$} \label{cond: frac-user}
\end{assumption}

When the items are assumed to be clustered, the items form $K$ item-clusters, each with size $M/K.$ Again, we assume items are indexed in such a way that item $m$ is in cluster $\lceil \frac{m}{M/K} \rceil=\lceil \frac{mK}{M} \rceil.$ We assume the items in the same cluster receive the same preference from the same user, i.e.,
$$b_{u m}= b_{u n}\quad \forall u \quad \hbox{if}\quad \left\lceil\frac{mK}{M}\right\rceil=\left\lceil\frac{nK}{M}\right\rceil.$$ We say the preference matrix $\bf B$ is {\em fractionally separable for items} if the following condition holds.
\begin{assumption} ({\bf Fractional Separability Condition for Items})  There exists a constant $0<\mu<1$ such that
\begin{eqnarray*}
&\sum_{u=1}^U {\bf 1}_{b_{um}=b_{un}} \leq \mu U &\quad \hbox{ if }\quad \left\lceil\frac{mK}{M}\right\rceil\not=\left\lceil\frac{nK}{M}\right\rceil\\
&\sum_{u=1}^U {\bf 1}_{b_{um}=b_{un}} = U &\quad \hbox{ if }\quad \left\lceil\frac{mK}{M}\right\rceil =\left\lceil\frac{nK}{M}\right\rceil
\end{eqnarray*}
In other words, for any pair of items $m$ and $n$ that are not in the same cluster, they receive the same preference from at most $\mu$ fraction of the users; and for any pair of items in the same cluster, they receive the same preference from all users.  \hfill{$\square$} \label{cond: frac-item}
\end{assumption}

The observed rating matrix $\bf R$ is assumed to be sparse and noisy because most users only rate a few items, and a user may give inconsistent ratings to the same items when being asked multiple times \cite{AmaPujOli_09}. The process of generating $\bf R$ is illustrated below, where $\bf B$ is first passed through a noisy channel to create $\tilde{{\bf R}},$ and then $\tilde{{\bf R}}$ is passed through an erasure channel to create $\bf R.$
\[
{\bf B}\xrightarrow{\hbox{a noisy channel}} \tilde{\bf R}  \xrightarrow{\hbox{an erasure channel}} {\bf R}.
\]

We assume the noisy channel satisfies a {\em biased rating} property under which a user is more likely to reveal the true preference.
\begin{assumption} ({\bf Biased Rating}) Given any $b_{um},$
\begin{eqnarray*}
\Pr\left(\tilde{r}_{um}=g\right)=\left\{
                                   \begin{array}{ll}
                                     p, & \hbox{ if } \quad g=b_{um} \\
                                     \frac{1-p}{G-1}, & \hbox{otherwise.}
                                   \end{array}
                                 \right.,
\end{eqnarray*}
where $p>\frac{1}{G}.$  So the probability that a user reveals the true preference is larger than the probability the user gives any other rating. \hfill{$\square$} \label{cond: biased}
\end{assumption}

Recall that $\bf R$ contains only a few entries of $\tilde{\bf R}.$ Let $r_{um}=\star$ if the entry is erased; and $r_{um}=\tilde{r}_{um}$ otherwise. We define two types of users: {\em information-rich users} who rate a large number of items and {\em information-sparse users} who rate only a few items. Specifically, the information-rich users and information-sparse users are defined as follows.
\begin{assumption} ({\bf Heterogeneous Users}) For an information-rich user $u,$ $$\Pr(r_{um}=\star)=1-\beta \quad\hbox{for all}\quad m;$$ and for an information-sparse user $v,$ $$\Pr(r_{vm}=\star)=1-\alpha\quad\hbox{ for all }\quad m.$$ In other words, an information-rich user rates $\beta M$ items on average; and an information-sparse user rates $\alpha M$ items on average. We further assume the erasures are independent across users and items, the number of information-rich users in each user-cluster is at least 2 and at most $\eta$ (a constant independent of $M$), $\alpha=o(\beta),$ and $\beta\leq \beta_{\max}<1.$ \hfill{$\square$}\label{cond: hert-user}
\end{assumption}

We further define two types of items: {\em information-rich items} that receive a large number of ratings and {\em information-sparse items} that receives only a few ratings. Specifically, the information-rich items and information-sparse items are defined in the following assumption.

\begin{assumption} ({\bf Heterogeneous Items}) For an information-rich item $m,$ $$\Pr(r_{um}=\star)=1-{\beta} \quad\hbox{for all}\quad u;$$ and for an information-sparse item $n,$ $$\Pr(r_{un}=\star)=1-{\alpha} \quad\hbox{ for all }\quad u.$$ In other words, an information-rich item receives ${\beta}U$ ratings on average; and an information-sparse item receives ${\alpha}U$ ratings on average.  We further assume the erasures are independent across users and items, the number of information-rich items in each item-cluster is at least 2 and at most $\eta$ (a constant independent of $M$), $\alpha=o(\beta),$ and $\beta\leq \beta_{\max}<1.$ \hfill{$\square$}\label{cond: hert-item}
\end{assumption}

{\bf Remark:} In real datasets, the number of ratings per user is small. To model this, we let $\alpha$ and $\beta$ be functions of $M$ which go to zero as $M\rightarrow\infty$. We assumed that $\alpha(M)=o(\beta(M))$  to model the information richness and sparsity. Also when the system has both information-rich users and information-rich items, we assume $\tilde{r}_{um}$ is erased with probability $\beta$ if either user $m$ is an information-rich user or item $m$ is an information-rich item; and $\tilde{r}_{um}$ is erased with probability $\alpha$ otherwise.

\subsection{Remarks on the conditions}
We present the conditions above in such a way that the notation in the analysis is simplified. Many of these conditions can be easily relaxed.  We next comment on these extensions, for which only minor modifications of the proofs are needed.
\begin{enumerate}[leftmargin=*]
\item Conditions \ref{cond: frac-user} and \ref{cond: frac-item}: These two conditions have been stated in a very general form and are easily satisfied. For example, note that if the blocks of $\bf B$ are chosen in some i.i.d. fashion, the conditions would hold asymptotically for large matrices with high probability. Furthermore, the constant $\mu$  can be different in the two conditions.

\item Condition \ref{cond: biased}: The noisy channel can be any channel that guarantees $$\Pr\left(\tilde{r}_{bm}=\tilde{r}_{vm}|b_{um}=b_{vm}\right)>\Pr\left(\tilde{r}_{bm}=\tilde{r}_{vm}|b_{um}\not=b_{vm}\right),$$ i.e.,  when two users have the same preference for a item, they are more likely to give the same rating than when they have different preferences for the item.

\item Conditions \ref{cond: hert-user} and \ref{cond: hert-item}: The upper bound $\eta$ can be a function of $M.$ The $\alpha$ and $\beta$ in the two conditions can also be different.

\item Cluster sizes: The cluster sizes can be different but of the same order. We also remark that in \cite{TomozeiMassoulie}, $K$ is assumed to be a constant, and in \cite{Dabeer}, PAF requires $K=O(\sqrt{M}).$ Our co-clustering algorithm, which will be presented in Section \ref{sec: alg}, works for $K=O(M/\log M).$

\end{enumerate}

Finally, we note that we do not require all the conditions to hold for the results in the paper. We will next present the results when a subset of these conditions hold and the results when all conditions hold.

\section{Main Results}
The focus of this paper is to derive the conditions under which the preference matrix $\bf B$ can be recovered from the observed rating matrix $\bf R,$ and develop high-performance algorithms. We assume it is a large-scale system and say an event occurs {\em asymptotically} if it occurs with probability one when $M \rightarrow \infty.$

We let $\Phi$ denote a matrix completion algorithm, and $\Phi({\bf R})$ denote the recovered matrix under algorithm $\Phi$ given observed rating matrix $\bf R.$ Further, we define $X_{\bf R}$ to be the number of observed ratings in $\bf R,$ i.e., $$X_{\bf R}=\sum_u\sum_m {\bf 1}_{r_{um}\not=\star},$$ where $\bf 1$ is the indicator function. {\em Our main results quantify the conditions required to asymptotically recover $\bf B$ from $\bf R$ in terms of the number of observed ratings $X_{\bf R}.$}

\subsection{Clustering for Recommendation}
We first assume the users are clustered and satisfy the fractional separability condition (\ref{cond: frac-user}).

\begin{thm}
Assume Conditions (\ref{cond: frac-user}), (\ref{cond: biased}) and (\ref{cond: hert-user}) hold. If $\alpha \leq \frac{ K}{U},$ then there exists a constant $\bar{U}$ such that for any matrix completion algorithm $\Phi$ and any $U\geq \bar{U},$ we can always find a rating matrix $\bf B$ such that $$\Pr(\Phi({\bf R})={\bf B}|{\bf B})\leq 1-\frac{\delta}{3},$$ where $\delta=(1-\beta_{\max})^{\eta} e^{-1.1}.$

Note that $$E[X_{\bf R}]\leq MK$$ implies that $\alpha \leq \frac{ K}{U}.$  So when the number of observed ratings is fewer than $MK,$  no matrix completion algorithm can recover all $\bf B$'s accurately.  \hfill{$\square$} \label{thm: l-clu}
\end{thm}

The proof of this theorem is presented in Appendix \ref{sec: proof-thm1}. The result is proved by showing that when  $\alpha \leq \frac{ K}{U}$ and $U$ is sufficiently large, if $$\Pr(\Phi({\bf R})={\bf B}|{\bf B})\geq 1-\frac{e^{-1.1}}{3}$$ for some $\bf B$, then we can construct $\hat{\bf B}$ such that $$\Pr(\Phi({\bf R})=\hat{\bf B}|\hat{\bf B})\leq 1-\frac{2e^{-1.1}}{3}.$$

\begin{thm}
Assume Conditions (\ref{cond: frac-user}), (\ref{cond: biased}), and (\ref{cond: hert-user}) hold. If $\alpha=\omega\left(\frac{K\log M}{M}\right)$ and $\alpha\beta =\omega\left(\frac{\log M}{M}\right),$ then there exists a matrix completion algorithm $\Phi$ such that given any $\epsilon>0,$ there exists $M_\epsilon$ such that
$$\Pr(\Phi({\bf R})={\bf B}|{\bf B})\geq 1-\epsilon$$ holds for any rating matrix $\bf B$ with at least $M_\epsilon$ items.

Note that $$E[X_{\bf R}]=\omega(MK\log M)$$ implies that $\alpha=\omega\left(\frac{K\log M}{M}\right).$ So there exists a matrix completion algorithm that can recover $\bf B$ asymptotically when $\alpha\beta =\omega\left(\frac{\log M}{M}\right)$ and number of observed ratings is $\omega(MK \log M).$ \hfill{$\square$}
\label{thm: fea-clu}
\end{thm}

The proof of this theorem can be found in Appendix \ref{sec: proof-thm2}. \textcolor{black}{Theorem \ref{thm: fea-clu} is established by presenting an algorithm which recovers the rating matrix asymptotically. This algorithm called User Clustering for Recommendation (UCR) is presented in Section \ref{sec: alg-clustering}. The algorithm is motivated by the PAF algorithm proposed in \cite{Dabeer}.} However, we made some key modifications to exploit the presence of information-rich users.  The key steps of our clustering algorithm are summarized below.
\begin{itemize}[leftmargin=*]
\item[(i)] User $u$ first compares her/his rating vector with other users, and selects an user who has the highest similarity to her/him (say user $v$). It can be proved that the selected user is an information-rich user in the same cluster.

\item[(ii)] Then the algorithm selects $U/K-2$ users according to their normalized similarity to user $v.$ It can be proved that these users are the users who are in the same cluster as user $v$ (so in the same cluster as user $u$).

\item[(iii)] For each item $m,$ the algorithm predicts $b_{um}$ via a majority vote among the selected users, including users $u$ and $v.$ The predicted rating is asymptotically correct.
\end{itemize}

\textcolor{black}{We note that theorems analogous to Theorems \ref{thm: l-clu} and \ref{thm: fea-clu} can be established for item clustering. The corresponding algorithm in the case will be called Item Clustering for Recommendation (ICR).}

\subsection{Co-Clustering for Recommendation}
We now assume both users and items are clustered and satisfy the fractionally separable conditions (\ref{cond: frac-user}) and (\ref{cond: frac-item}).
\begin{thm}
Assume Conditions (\ref{cond: frac-user}), (\ref{cond: frac-item}), (\ref{cond: biased}), (\ref{cond: hert-user}), and (\ref{cond: hert-item}) hold. If $\alpha \leq \frac{K^2}{MU}$ and $\beta\leq \frac{K}{\eta(M+U)-\eta^2K},$  then there exist a constant $\bar{M}$ such that for any matrix completion algorithm $\Phi$ and any $M\geq \bar{M},$ we can always find a rating matrix $\bf B$ such that $$\Pr(\Phi({\bf R})={\bf B}|{\bf B})\leq 1-\frac{\delta}{3},$$ where $\delta=e^{-2.2}.$

Note that $$E[X_{\bf R}]\leq K^2$$ implies that $\alpha \leq \frac{K^2}{UM}$ and $\beta\leq \frac{ K}{\eta(M+U)-\eta^2 K}.$ So when the number of observed ratings is fewer than $K^2,$  no matrix completion algorithm can recover all $\bf B'$s accurately. \hfill{$\square$} \label{thm: l-co-clu}
\end{thm}

The detailed proof is presented in Appendix \ref{sec: proof-thm3}.

\begin{thm}
Assume Conditions (\ref{cond: frac-user}), (\ref{cond: frac-item}), (\ref{cond: biased}), (\ref{cond: hert-user}), and (\ref{cond: hert-item}) hold. Further, assume the following conditions hold:
\begin{itemize}
\item[(i)] $\alpha =\omega\left(\frac{K^2\log M}{M^2}\right)$ or $\beta=\omega\left(\frac{K\log M}{M}\right);$ and
\item[(ii)] $\alpha\beta =\omega\left(\frac{\log M}{M}\right)$ or $\beta^2 =\omega\left(\frac{\log M}{K}\right).$
\end{itemize} Then there exists a matrix completion algorithm $\Phi$ such that given any $\epsilon>0,$ there exists $M_\epsilon$ such that
$$\Pr(\Phi({\bf R})={\bf B}|{\bf B})\geq 1-\epsilon$$ holds for any rating matrix $\bf B$ with at least $M_\epsilon$ items.

Note that $$E[X_{\bf R}]=\omega(K^2\log M)$$ implies condition (i), so $\bf B$ can be asymptotically recovered from $\bf R$ when the number of observed ratings is $\omega(K^2\log M)$ and condition (ii) holds. \hfill{$\square$} \label{thm: fea-co-clu}
\end{thm}
Theorem \ref{thm: fea-co-clu} is proved by showing that there exists a co-clustering algorithm that clusters both users and items. Then the preference $b_{um}$ is recovered by a majority vote among all observed $r_{vn}$s for all $v$ in the same user-cluster as $u$ and all  $n$ in the same item-cluster as $m.$ \textcolor{black}{We have relegated the proof of Theorem \ref{thm: fea-co-clu} to the appendix since the main ideas behind the proof are similar to the proof of Theorem \ref{thm: fea-clu}.} The detailed description of the co-clustering algorithm, named Co-Clustering for Recommendation (CoR), is presented in Section \ref{sec: alg-co-clustering}.

We summarize our main results, and compare them with corresponding results for matrix completion, in Table \ref{tab: main}. 

\begin{table*}[hbt]
\centering
\begin{tabular}{|c|c|c|c|}
  \hline
    & Clustering & Co-Clustering & Matrix Completion with Rank $K$\\
\hline
  Necessary &  $E[X_{\bf R}]=\gamma MK \log M$  & $E[X_{\bf R}]=\gamma K^2 $ & $E[X_{\bf R}]=\gamma MK\log M$ \cite{CandesTao}\\
\hline
  Sufficient  & $E[X_{\bf R}]=\omega(MK\log M),$ and  & $ E[X_{\bf R}]=\omega(K^2\log M),$ and & $E[X_{\bf R}]=\Omega(MK\log^2 M)$ \cite{Recht}\\
    &     $\alpha\beta M=\omega(\log M)$ & $\alpha\beta M=\omega(\log M)$ or $\beta^2K=\omega(\log M)$ & \\
  \hline
\end{tabular}
\caption{A summary of the main results}\label{tab: main}
\end{table*}

\section{Algorithms}
\label{sec: alg}

\subsection{Clustering for Recommendation}
\label{sec: alg-clustering}
Before presenting the algorithms, we first introduce the notions of co-rating and similarity. Given users $u$ and $v,$
the co-rating of the two users is defined to be the number of items they both rate:
\[
\varphi_{u,v}=\sum_{m=1}^M {\bf 1}_{r_{vm}\not=\star, r_{um}\not=\star};
\]
and the similarity of the two users is defined to be the number of items they rate the same  minus the number of items they rate differently:
\begin{eqnarray*}
\sigma_{u,v}&=& \sum_{m=1}^M {\bf 1}_{r_{um}=r_{vm}\not=\star}-\sum_{m=1}^M {\bf 1}_{r_{um}\not=r_{vm}, r_{vm}\not=\star, r_{um}\not=\star}\\
&=&2\sum_{m=1}^M {\bf 1}_{r_{um}=r_{vm}\not=\star}-\varphi_{u,v}.
\end{eqnarray*}
We further define the normalized similarity of two users to be
\[
\tilde{\sigma}_{u,v}=\frac{\sigma_{u,v}}{\varphi_{u,v}}=\frac{2\sum_{m=1}^M {\bf 1}_{r_{um}=r_{vm}\not=\star}}{\varphi_{u,v}}-1.
\]
Similarly, we can define the co-rating,  similarity and normalized similarity of two items:
\begin{eqnarray*}
\varphi_{m,n}&=&\sum_{u=1}^U  {\bf 1}_{r_{u m}\not=\star, r_{u n}\not=\star}\\
\sigma_{m,n}&=&2\sum_{u=1}^U {\bf 1}_{r_{um}=r_{un}\not=\star}-\varphi_{u,v}\\
\tilde{\sigma}_{m,n}&=&\frac{2\sum_{u=1}^U {\bf 1}_{r_{um}=r_{un}\not=\star}}{\varphi_{m,n}}-1.
\end{eqnarray*}

When users are clustered, the following algorithm exploits the existence of both the cluster structure and information-rich entities to recover the matrix.

\noindent\hrulefill

\noindent{\bf User Clustering for Recommendation (UCR)}
\begin{enumerate}[leftmargin=*]
\item[(i)] For user $u,$ the algorithm selects a user $v$ who has the highest similarity to user $u,$ i.e.,  $$v \in \arg\max_{w\not= u}  \sigma_{u,w}.$$

\item[(ii)] The algorithm then selects $\frac{U}{K}-2$ users in a descending order according to their normalized similarity to user $v.$ Define ${\cal F}_u$ to be the set of the selected  $\frac{U}{K}-2$ users, user $v$ and user $u.$

\item[(iii)] For each item $m,$ the preference $b_{w m}$ for $w\in{\cal F}_u$ is determined by a majority vote among the users in ${\cal F}_u,$ i.e., $$b_{w m}=\arg\max_g \sum_{v\in {\cal F}_u} {\bf 1}_{r_{vm}=g}.$$
\end{enumerate}

\noindent\hrulefill

When items are clustered, a similar algorithm that clusters items can be used to recover the matrix. The algorithm is named Item Clustering for Recommendation (ICR), and is presented below.

\noindent\hrulefill

\noindent{\bf Item Clustering for Recommendation (ICR)}
\begin{itemize}[leftmargin=*]
\item[(i)] For item $m,$ the algorithm selects an item $n$ that has the highest similarity to item $m.$

\item[(ii)] The algorithm then selects $\frac{M}{K}-2$ items in a descending order according to their normalized similarity to item $n.$ Define ${\cal N}_m$ to be the set of the selected  $\frac{M}{K}-2$ items, item $m$ and item $n.$

\item[(iii)] For each user $u,$ the preference $b_{ul}$ for $l\in{\cal N}_m$ is determined by a majority vote among the items in ${\cal F}_m,$ i.e., $$b_{ul}=\arg\max_g \sum_{n\in {\cal F}_m} {\bf 1}_{r_{un}=g}.$$
\end{itemize}

\noindent\hrulefill

\subsection{Co-Clustering for Recommendation}
\label{sec: alg-co-clustering}

When both users and items are clustered, we propose the following co-clustering algorithm for recovering $\bf B.$ For given a (user, item) pair, the algorithm identifies the corresponding user-cluster and item-cluster, and then uses a majority vote within the $\frac{U}{K}\times \frac{M}{K}$ block to recover $b_{um}.$

\noindent\hrulefill

\noindent{\bf Co-Clustering for Recommendation (CoR)}
\begin{itemize}[leftmargin=*]
\item[(i)] Given a (user, item) pair $(u,m),$ the algorithm calls steps (i) and (ii) of UCR to obtain a set of users ${\cal F}_u;$ and calls steps (i) and (ii) of ICR to obtain a set of items ${\cal N}_m.$

\item[(ii)] For each item $n\in {\cal N}_m$ and each user $v\in {\cal F}_u,$  $b_{vn}$ is decided by a majority vote among $r_{wl}$'s for $w\in{\cal F}_u$ and $l\in{\cal N}_m,$ i.e., $$b_{vn}=\arg\max_g \sum_{w\in {\cal F}_u} \sum_{l \in{\cal N}_m}{\bf 1}_{r_{wl}=g}.$$
\end{itemize}

\noindent\hrulefill

\subsection{Hybrid Algorithms}

As we mentioned in the introduction, in practice, it is hard to verify whether the cluster assumptions hold for the preference matrix since the matrix is unknown except for a few noisy entries. Even if the cluster assumptions hold, it is hard to check whether every user-cluster (item-cluster) contains information-rich users (items). When a user-cluster does not contain an information-rich user, UCR is likely to pick an information-rich user from another cluster in step (i) and results in selecting users from a different cluster in step (ii). So when a user-cluster has no information-rich user, it is better to skip step (i) and select users according to their normalized similarity to user $u;$ or just use PAF.

Since it is impossible to check whether a user-cluster has information-rich users or not, we propose the following hybrid algorithms, which combine three different approaches. We note that the hybrid algorithms are heuristics motivated by the theory developed earlier. Using the hybrid user-clustering for recommendation as an example, it combines the following three approaches: (i) first find an information-rich user and then use users' similarities to the selected information rich-user to find the corresponding user-cluster, (ii) directly use other users' similarities to user $u$ to find the corresponding user-cluster, and (iii) directly use other users' {\em normalized} similarities to user $u$ to find the corresponding user-cluster. After identifying the three possible clusters, the algorithm aggregates all the users in one cluster into a super-user, and computes the similarity between the super-user and user $u.$ The super-user with the highest similarity is used to recover the ratings of user $u.$  We further modify the definition of normalized similarity because this new normalized similarity works better than the original one in the experiments with real data sets:
\[
\tilde{\sigma}_{u,v}=\frac{\sigma_{u,v}}{\sqrt{\sum_{m=1}^M {\bf 1}_{r_{vm}\not=\star}}}.
\]

We next present the detailed description of the Hybrid User-Clustering for Recommendation (HUCR), the Hybrid Item-Clustering for Recommendation (HICR) and the Hybrid Co-Clustering for Recommendation (HCoR).  

\noindent\hrulefill

\noindent{\bf Hybrid User-Clustering for Recommendation (HUCR)}

\begin{itemize}[leftmargin=*]
\item[(i)] For each user $u,$ the algorithm calls steps (i) and (ii) of UCR to obtain a set of $T-1$ users, not including user $u.$ Denote the set by ${\cal F}^1_u.$

\item[(ii)] The algorithm selects $T-1$ users in a descending order according to their modified normalized similarity to users $u.$ Denote the set by ${\cal F}^2_u.$

\item[(iii)] The algorithm selects $T-1$ users in a descending order according to their similarity to user $u.$ Denote the set by ${\cal F}^3_u.$

\item[(iv)] The algorithm defines three super-users $s_z$ ($z=1, 2, 3$) such that the ratings of $s_z$ is determined by a majority vote among the users in ${\cal F}^z_u,$ i.e.,
$$r_{s_z,m}=\arg\max_g \sum_{v\in {\cal F}^z_u} {\bf 1}_{r_{v,m}=g}.$$

\item[(v)] The algorithm selects the super-user who has the highest similarity to user $u.$ Without the loss of generality, assume super-user $s_1$ is selected, then the ratings of user $u$ is given by $b_{u,m}=r_{s_1,m}.$
\end{itemize}

\noindent\hrulefill

{\bf Remark:} Here we could use the cluster size as $T.$ But we use a new variable $T$ here to emphasize the fact that the hybrid algorithms are designed for real datasets where we may not know the cluster size. In practice, we estimate $T$ by noting that the similarity score undergoes a phase transition when too many similar users are selected.

{\bf Remark:} The algorithm uses similarity in step (v). In fact, the three super-users will be information-rich users, so there is no significant difference between using similarity and using normalized similarity.

\noindent\hrulefill

\noindent{\bf Hybrid Item-Clustering for Recommendation (HICR)}

\begin{itemize}[leftmargin=*]
\item[(i)] For each item $m,$ the algorithm calls steps (i) and (ii) of ICR to obtain a set of $T-1$ items, not including item $m.$ Denote the set by ${\cal N}^1_m.$

\item[(ii)] The algorithm selects $T-1$ items in a descending order according to their modified normalized similarity to item $m.$ Denote the set by ${\cal N}^2_m.$

\item[(iii)] The algorithm selects $T-1$ items in a descending order according to their similarity to item $m.$ Denote the set by ${\cal N}^3_m.$

\item[(iv)] The algorithm defines three super-items $s_z$ ($z=1, 2, 3$) such that the ratings of $s_z$ is determined by a majority voting among the items in ${\cal N}^z_m,$ i.e.,
$$r_{u, s_z}=\arg\max_g \sum_{n\in {\cal N}^z_m} {\bf 1}_{r_{u,n}=g}.$$

\item[(v)] The algorithm selects the super-item that has the highest similarity to item $m.$ Without the loss of generality, assume super-item $s_1$ is selected, then the ratings of item $m$ is given by $b_{u,m}=r_{u, s_1}.$
\end{itemize}

\noindent\hrulefill

\noindent\hrulefill

\noindent{\bf Hybrid Co-Clustering for Recommendation (HCoR)}
\begin{itemize}[leftmargin=*]
\item[(i)] For each (user, item) pair $(u, m),$ the algorithm calls steps (i)-(iv) of HUCR to obtain the set of $T-1$ users forming the super-user who has the highest similarity to user $u,$ denoted by ${\cal F}_u;$ and steps (i)-(iv) of HICR to obtain the set of $T-1$ items forming the super-item that has the highest similarity to item $m,$ denoted by ${\cal N}_m.$

\item[(ii)] The rating $r_{mu}$ is determined as follows:
\begin{eqnarray*}
r_{mu}&=&\arg\max_g \sum_{v\in{\cal F}_u} {\bf 1}_{r_{vm}=g}+\sum_{n\in{\cal N}_m} {\bf 1}_{r_{un}=g}\\
&&+\sqrt{\sum_{v\in {\cal F}_u, n\in{\cal N}_m} {\bf 1}_{r_{vn}=g}}.
\end{eqnarray*}
\end{itemize}

\noindent\hrulefill

\begin{figure}[htb]
\centering
\includegraphics[width=1.8in]{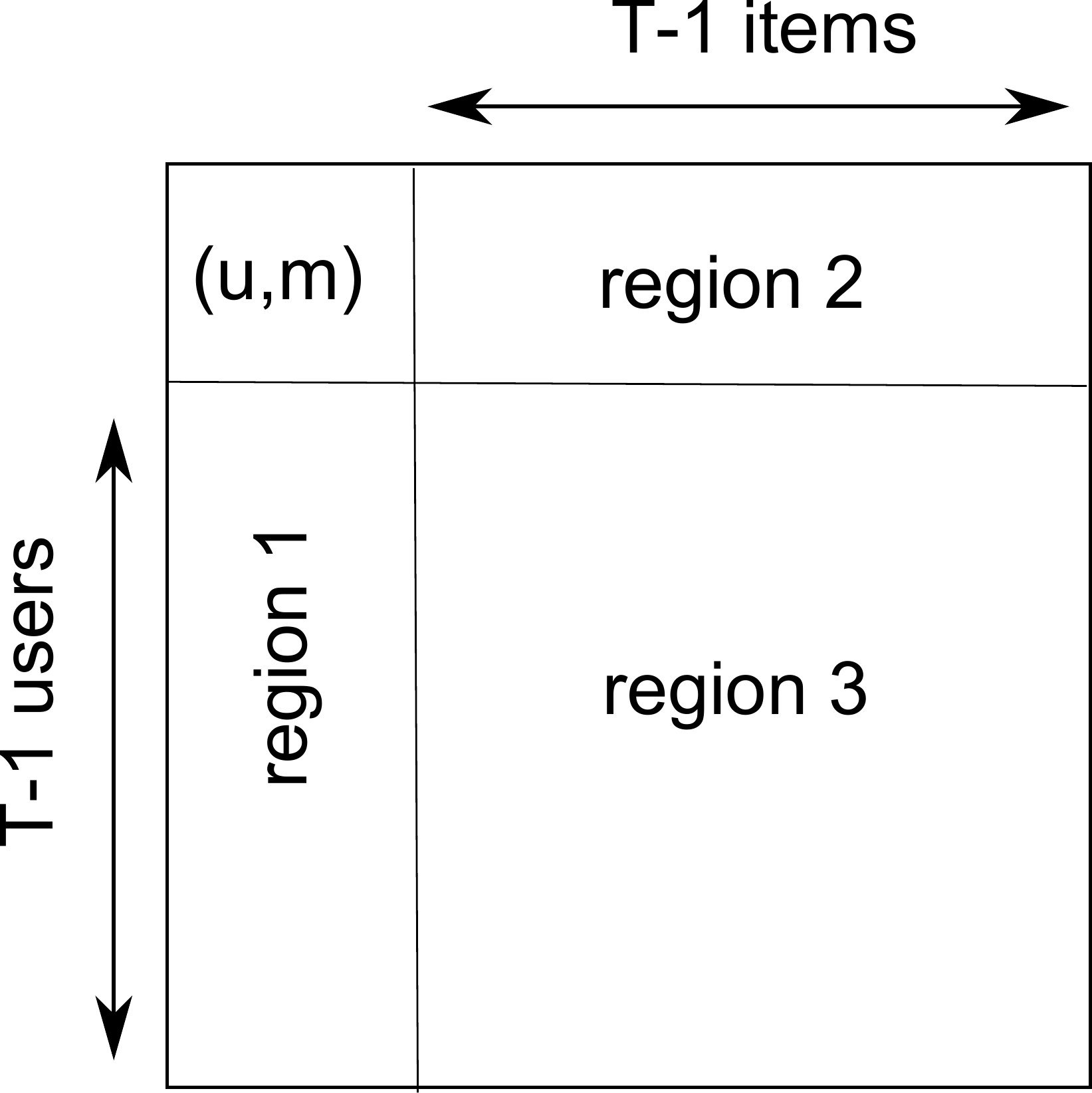}
\caption{The block ratings associated with (user, item) pair $(u, m)$}
\label{fig: HCoR-mv}
\end{figure}

{\bf Remark:} We used modified majority voting in step (ii) of HCoR because of the following reason: After step (i), the algorithm identifies $T-1$ users and $T-1$ items as shown in Figure \ref{fig: HCoR-mv}. The ratings in region 1 (i.e.,  $r_{vm}$ for $v\in{\cal F}_u$) are the ratings given to item $m$ and the ratings in region 2 (i.e., $r_{un}$ for $n\in{\cal N}_m$) are the ratings given by user $u.$  The ratings in region 3 (i.e., $r_{vn}$ for $v\in {\cal F}_u$ and $n\in{\cal N}_m$) are the ratings given by the users similar to user $u$ to the items similar to item $m.$ In our experiments with real datasets, we found that the ratings in region 1 and 2 are more important in predicting $b_{um}$ than the ratings in region 3. Since we have $(T-1)\times(T-1)$ entries in region 3 but only $T-1$ entries in region 1 or 2, so we use the square-root of the votes from region 3 to reduce its weight in the final decision.

\section{Performance Evaluation}
\label{sec: per}
In this section, we evaluate the performance of our hybrid clustering and co-clustering algorithms and compare them with PAF \cite{Dabeer} and AM \cite{SheWenZha_12}.  We tested the algorithms using both the MoiveLens dataset \cite{MovieLensData} and Netflix dataset \cite{NetflixData}.  Our main goal is to recommend to each user only those movies that are of most interest to that user. In other words, we only want to recommend movies to a user that we believe would have been rate highly by that user. Therefore, we quantize the ratings in both datasets so that movies which received a rating greater than $3.5$ are reclassified as $+1$ and movies which received a rating of $3.5$ or below are reclassified as $-1.$ This binary quantization is also necessary to make a fair comparison with the results in \cite{Dabeer} since the algorithm there only works for binary ratings. For both datasets, we hide 70\% of the ratings. So 30\% of the ratings were used for training (or predicting); and 70\% of the ratings were used for testing the performance.  The following three performance metrics are used in the comparison:
\begin{enumerate}[leftmargin=*]
\item {\bf Accuracy at the top:}  This terminology was introduced in \cite{SteCorMeh_12} which in our context means the accuracy with which we identify the movies of most interest to each user. In our model, instead of computing accuracy, we compute the error rate (the fraction of ratings that are not correctly recovered) when we recommend a few top items to each user, which they may like the most. But we continue to use the term ``accuracy at the top" to be consistent with prior literature. Since the goal of a recommendation system is indeed to recommend a few items that a user may like, instead of recovering a user's preference to all items, we view this performance metric as the most important one among the three metrics we consider in this paper.

    In HCoR and PAF, the top-items were selected based on majority voting within each cluster, and in AM, the top-items were selected based on the recovered value. To make a fair comparison among the five algorithms, we restricted the algorithms to only recommend those items whose ratings were given in the dataset but hidden for testing.

\item {\bf Accuracy for information-sparse users:} In real datasets, a majority of users only rate a few items. For example, in the MovieLens dataset, more than 73.69\% users only rated fewer than 200 movies, and their ratings only consist of 34.32\% of the total ratings. The accuracy for information-sparse users measures the error rate for these information-sparse users who form the majority of the user population. Note the overall error rate is biased towards those information-rich users who rated a lot of movies (since they account for 65.68\% of the ratings in the MovieLens data, for example).

\item {\bf Overall accuracy:} The overall error rate when recovering the rating matrix. We include this metric for completeness.
\end{enumerate}

Before presenting the detailed results, for the convenience of the reader, we summarize the key observations:
\begin{enumerate}[leftmargin=*]
\item In all of the datasets that we have studied, our co-clustering algorithm HCoR consistently performs better than the previously-known PAF and AM algorithms. When noise is added to the datasets, the performance difference between our algorithm and the AM algorithm increases even more.

\item If the goal is to recommend a small number of movies to each user (i.e., concerning accuracy at the top), then even the user clustering algorithm HUCR performs better than PAF and AM, but worse than HCoR.

\item We have not shown other studies that we have conducted on the datasets due to space limitations, but we briefly mention them here. If the goal is to recommend a few users for each item (for example, a new item may be targeted to a few users), then item clustering performs better than PAF and AM, but HCoR still continues to perform the best. Also, simple clustering techniques such as spectral clustering, following by a majority vote within clusters, do not work as well as any of the other algorithms.
\end{enumerate}
In the following subsections, we present substantial experimental studies that support our key observations 1 and 2.
\subsection{MovieLens Dataset without Noise}
We conducted experiments on the MovieLens dataset \cite{MovieLensData}, which has 3,952 movies, 6,040 users and 1,000,209 ratings. \textcolor{black}{So users rate about 4\% of the movies in the MovieLens dataset.}

\begin{figure}[htb]
\centering
\includegraphics[width=2.8in]{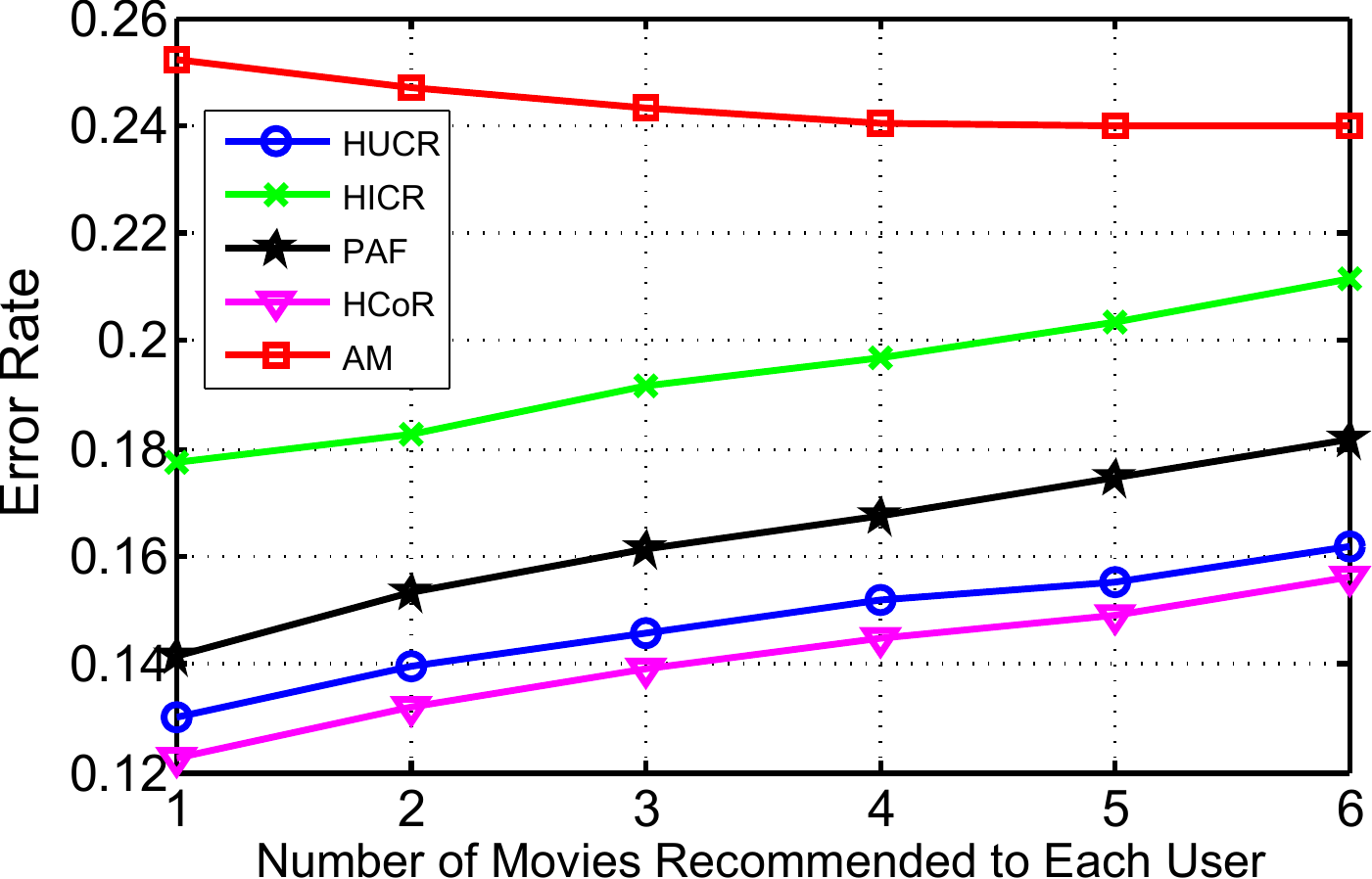}
\caption{Accuracy at the top for the MovieLens dataset. The figure shows the error rates when we recommend $x$ movies to each user.}
\label{fig: 30-70-item-to-user}
\end{figure}

We first evaluated the accuracy at the top. Figure \ref{fig: 30-70-item-to-user} shows the error rate when we recommended $x$ movies to each user for $x=1, 2, \cdots, 6.$ We can see that HCoR performs better than all other algorithms, and HUCR has a similar error rate as HCoR. In particular, when one movie is recommended to each user, HCoR has an error rate of 12.27\% while AM has an error rate of 25.22\% and PAF has an error rate of 14\%.

\begin{figure}[htb]
\centering
\includegraphics[width=2.8in]{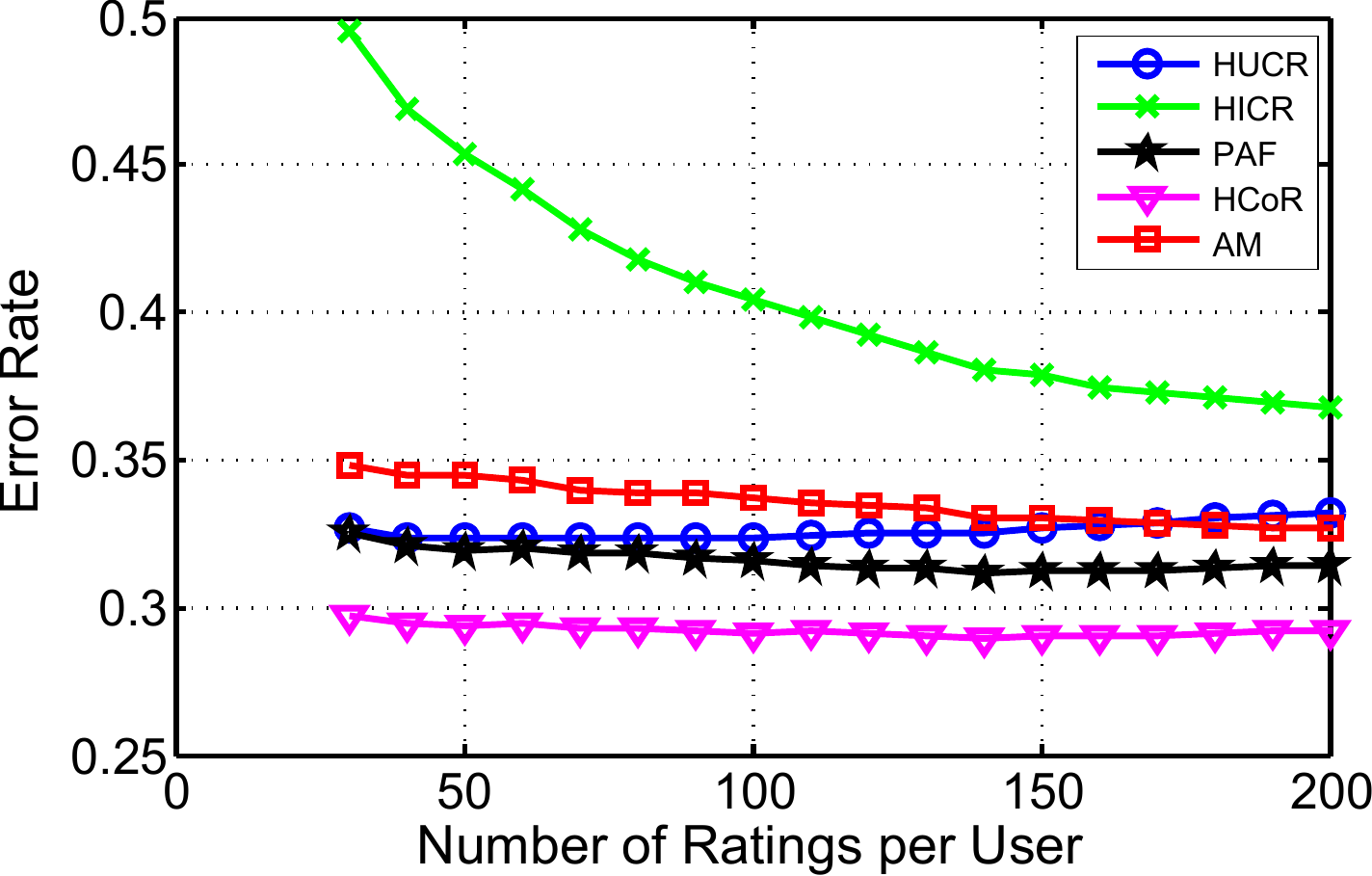}
\caption{Accuracy for information-sparse users for the MovieLens dataset. The figure shows the error rates of users who rate different numbers of movies.}
\label{fig: 30-70-sparse-u}
\end{figure}

We then evaluated the accuracy for information-sparse users. Figure \ref{fig: 30-70-sparse-u} shows the error rate for the users who rate between less than $x$ movies, for $x=30, 40, \cdots, 200.$ We can see from the figure that HCoR has the lowest error rate. For example, for users who rated less than 30 movies, HCoR has an error rate of 29.72\% while AM has an error rate of 34.81\%.

For completeness, we also summarize the overall error rates in Table \ref{tab: er-ml}. HCoR has the lowest error rate.
\begin{table}[hbt]
\centering
\begin{tabular}{|c|c|c|c|c|}
  \hline
  HUCR  & HICR & HCoR & PAF & AM \\
\hline
 34.69\%& 32.87\%& 29.2\% & 32.07\% & 30.62\%\\
 \hline
\end{tabular}
\caption{The overall error rates for the MovieLens dataset} \label{tab: er-ml}
\end{table}

{\bf Remark:} It is possible that some ratings cannot be obtained under HUCR, HICR, HCoR and PAF, e.g., $b_{um}$ cannot be obtained when none of selected users in step (ii) of HUCR rated movie $m.$ When it occurs, we counted it as an error. So we made very conservative calculations in computing the error rates for HUCR, HICR, HCoR and PAF.

\subsection{Netflix Dataset without Noise}
We conducted experiments on the Netflix dataset \cite{NetflixData}, which has 17,770 movies, 480,189 users and 100,480,507 ratings. We used all movies but randomly selected 10,000 users, which gives 2,097,444 ratings for our experiment. The reason that we selected 10,000 users is that, otherwise, the dataset is too large to handle without the use of special purpose computers. In particular, it is not clear how one would implement that AM algorithm on the full dataset since the first step of that algorithm requires one to perform an SVD which is not possible using a computer with 8G RAM, 2.5 Ghz processor that we used.

\begin{figure}[htb]
\centering
\includegraphics[width=2.8in]{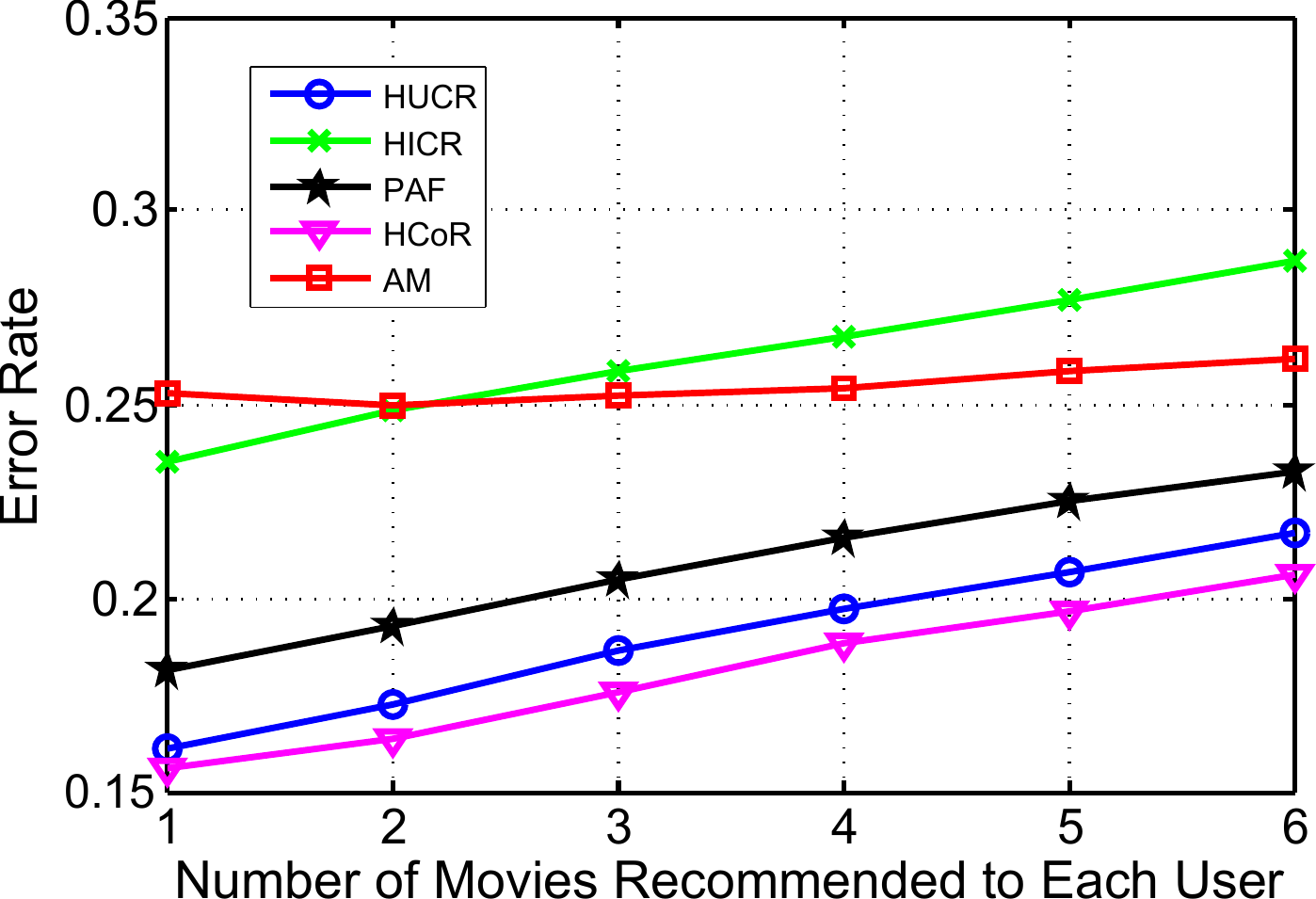}
\caption{Accuracy at the top for the Netflix dataset. The figure shows the error rates when we recommend $x$ movies to each user.}
\label{fig: nf-30-70-item-to-user}
\end{figure}

Figure \ref{fig: nf-30-70-item-to-user} shows the accuracy at the top, i.e., the error rate when we recommended $x$ movies to each user for $x=1, 2, \cdots, 6.$ We can see that HCoR performs better than all other algorithms, and HUCR has a similar error rate as HCoR. When one movie is recommended to each user, HCoR has an error rate of 15.58\% while AM has an error rate of 25.29\%.

\begin{figure}[htb]
\centering
\includegraphics[width=2.8in]{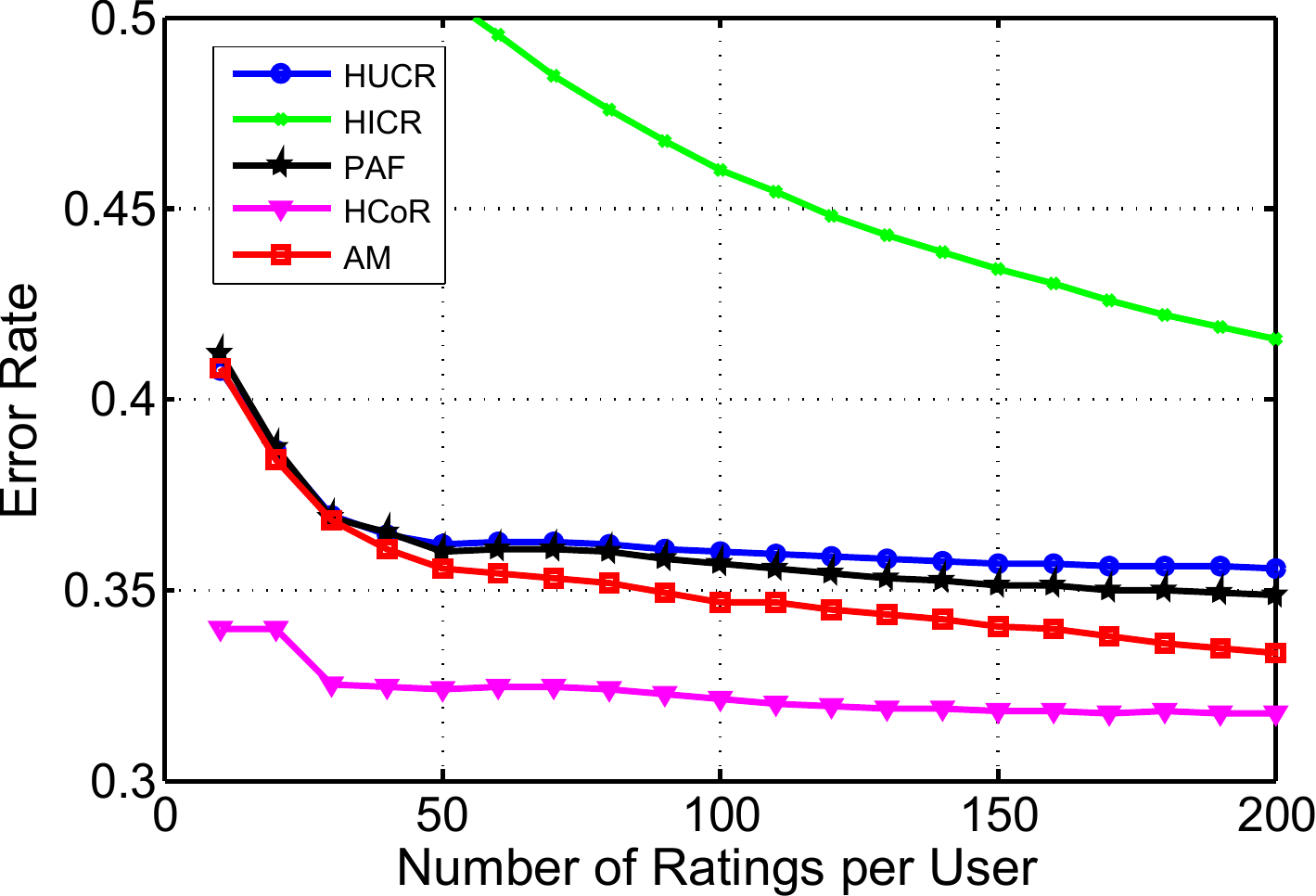}
\caption{Accuracy for information-sparse users for the Netflix dataset. The figure shows the error rates of users who rate different numbers of movies.}
\label{fig: nf-30-70-sparse-u}
\end{figure}

We then evaluated the accuracy for information-sparse users. Figure \ref{fig: nf-30-70-sparse-u} shows the error rate for the users who rate less than $x$ items, for $x=10, 20, 30, \cdots, 200.$ Among the 10,000 randomly selected users, 70\% of them rated no more than 200 movies. We can see from the figure that HCoR has the lowest error rate. In particular, for the users who rate less than 10 items, HCoR has an error rate of 33.95\% while the error rate of AM is 40.81\%.

Table \ref{tab: er-nf} summarizes the overall error rates. AM and HCoR have the lowest overall error rate.
\begin{table}[hbt]
\centering
\begin{tabular}{|c|c|c|c|c|}
  \hline
  HUCR  & HICR & HCoR & PAF & AM \\
\hline
 36.49\%& 34.72\%& 31.11\% & 35.51\% & 31.11\%\\
 \hline
\end{tabular}
\caption{The overall error rates for the Netflix dataset} \label{tab: er-nf}
\end{table}

\subsection{MovieLens Dataset with Noise}
\begin{figure}[htb]
\centering
\includegraphics[width=2.8in]{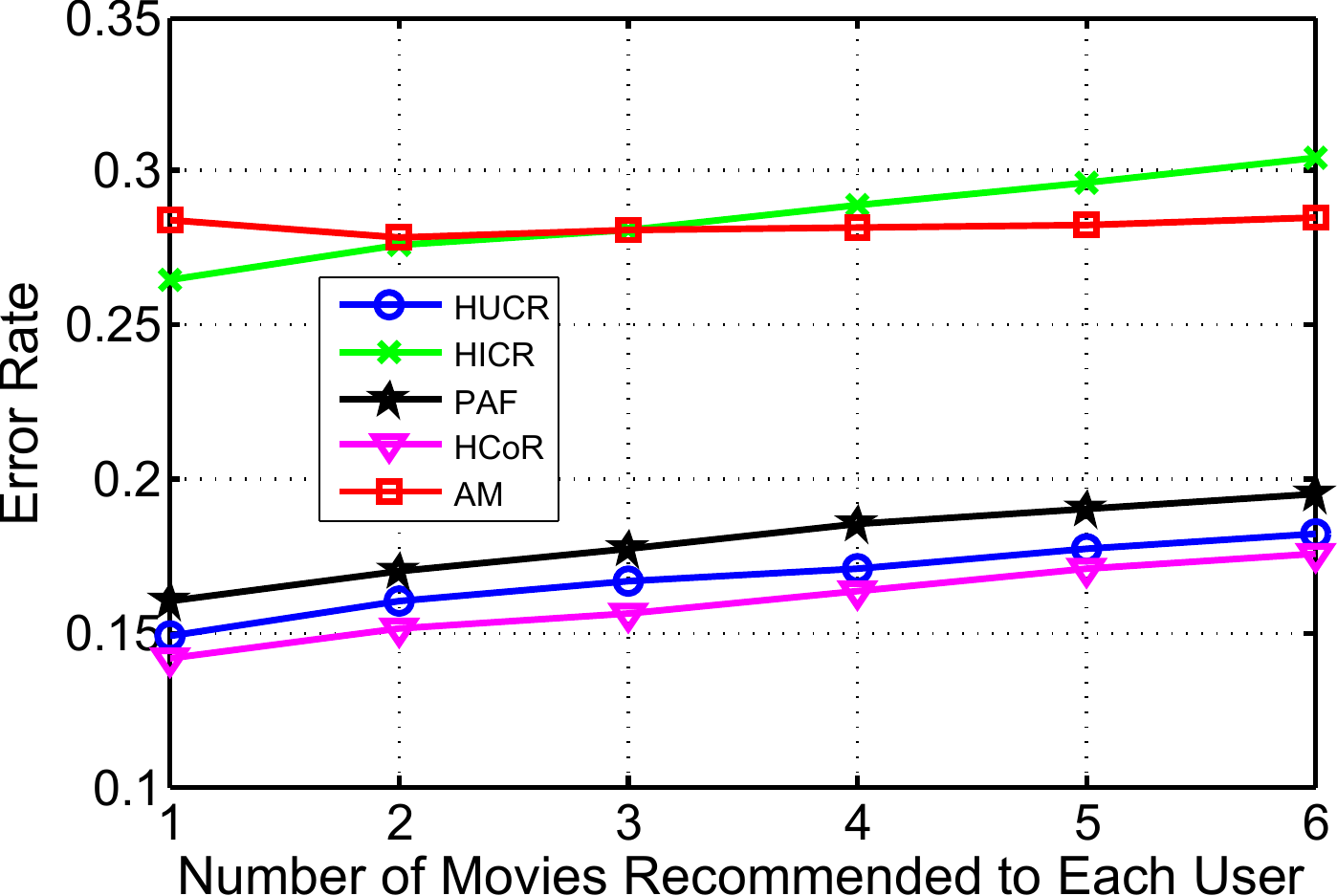}
\caption{Accuracy at the top for the MovieLens data with noise. The figure shows the error rates when we recommend $x$ movies to each user.}
\label{fig: 30-70-item-to-user-2n}
\end{figure}

Our theoretical results suggested that our clustering and co-clustering algorithms are robust to noise. In this set of experiments, we independently flipped each un-hidden rating with probability $0.2,$ and then evaluated the performance of our clustering and co-clustering algorithms.  The result for the accuracy at the top is shown in Figure \ref{fig: 30-70-item-to-user-2n}. We can see that HCoR performs better than all other algorithms. When one movie is recommended to each user, HCoR has an error rate of 14.19\% while AM has an error rate of 28.41\%. Comparing to the noise-free case, the error rate of HCoR increases by 1.92\%, and the error rate of AM increases by 3.19\%.

\begin{figure}[htb]
\centering
\includegraphics[width=2.8in]{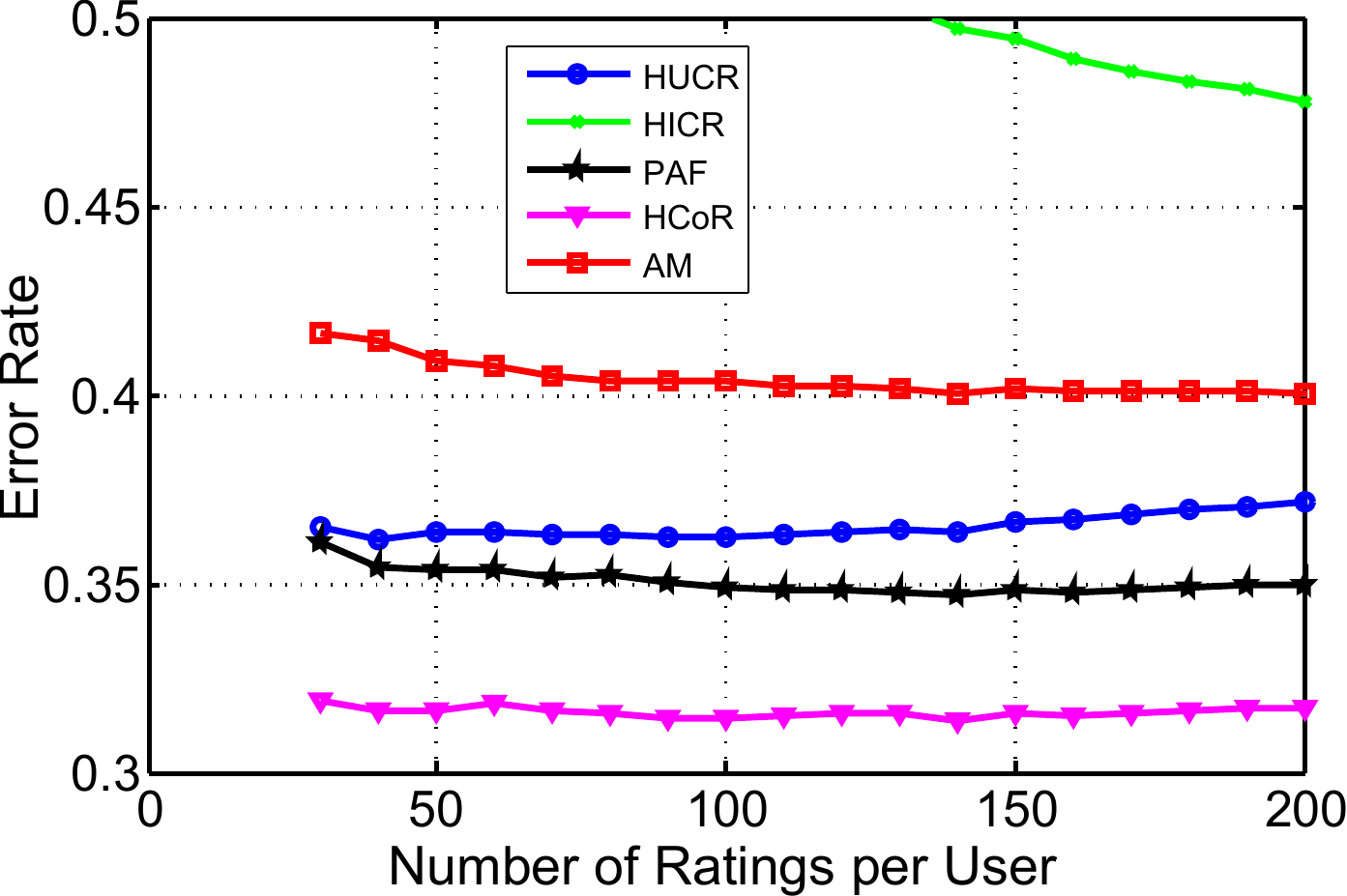}
\caption{Accuracy for information-sparse users for the MovieLens data with noise. The figure shows the error rates of users who rate different numbers of movies.}
\label{fig: 30-70-sparse-u-0.2n}
\end{figure}

The results for the accuracy for information-sparse users are presented in Figure \ref{fig: 30-70-sparse-u-0.2n}. HCoR has the lowest error rate. For users who rated less than 30 movies, HCoR has an error rate of 31.87\% while AM has an error rate of 41.67\%. Comparing to the noise-free case, the error rate of HCoR increases only by 2.15\%, but the error rate of AM increases by 6.86\%.

For completeness, we also summarize the overall error rates in Table \ref{tab: er-ml-2n}. HCoR has the lowest error rate, and AM has a significantly higher error rate in this case. Note that HCoR and AM have similar overall error rates in the noise-free case.  From this set of experiments, we can see that HCoR is more robust to noise than AM.
\begin{table}[hbt]
\centering
\begin{tabular}{|c|c|c|c|c|}
  \hline
  HUCR  & HICR & HCoR & PAF & AM \\
\hline
 40.95\%& 41.95\%& 32.55\% & 35.64\% & 38.46\%\\
 \hline
\end{tabular}
\caption{The overall error rates of the MovieLens dataset with noise} \label{tab: er-ml-2n}
\end{table}

\subsection{Netflix Dataset with Noise}
\begin{figure}[htb]
\centering
\includegraphics[width=2.8in]{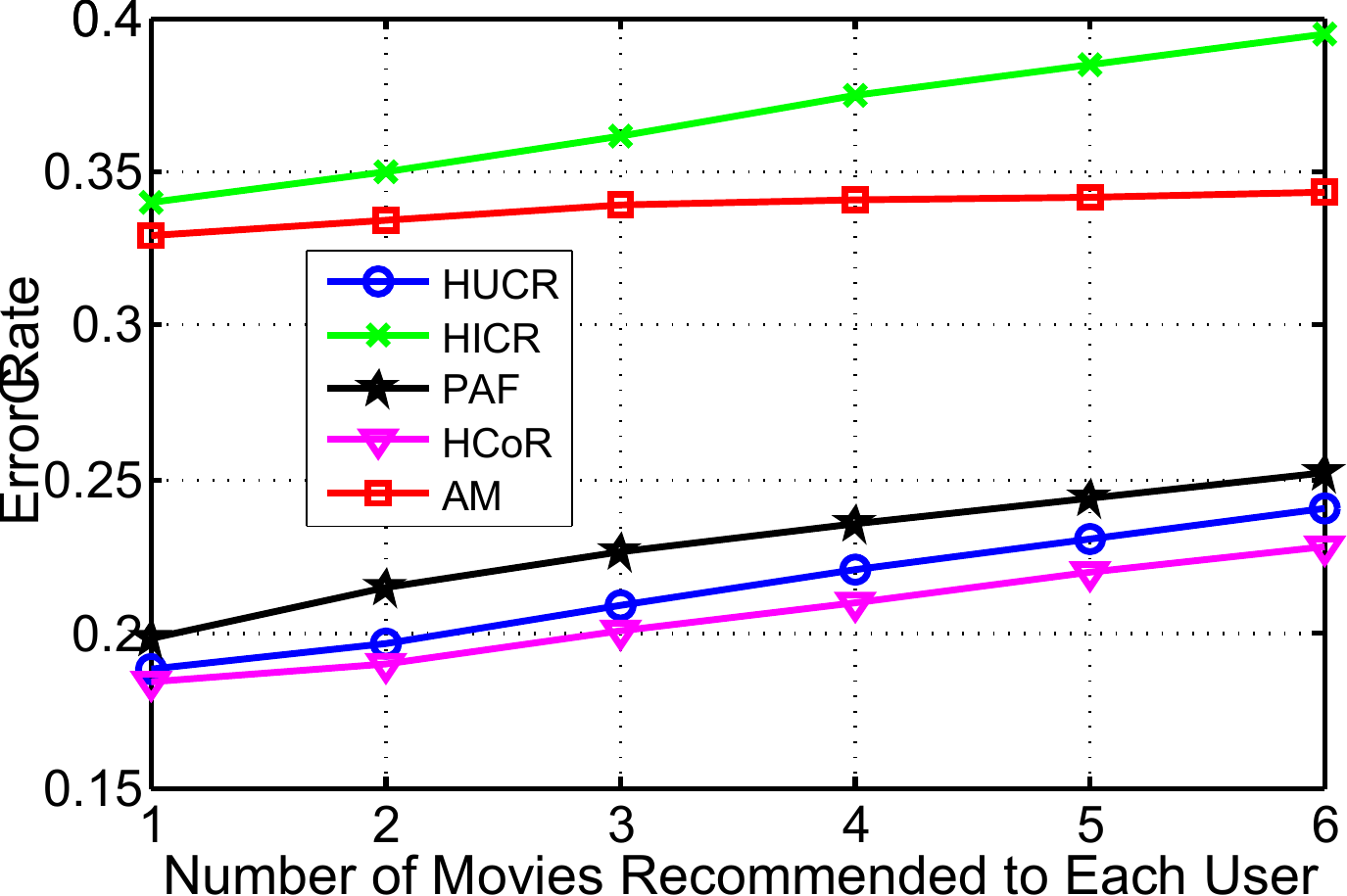}
\caption{Accuracy at the top for the Netflix data with noise. The figure shows the error rates when we recommend $x$ movies to each user.}
\label{fig: nf-30-70-item-to-user-2n}
\end{figure}

In this set of experiments, we flipped each un-hidden rating of the Netflix dataset \cite{NetflixData} with probability $0.2.$ Figure \ref{fig: nf-30-70-item-to-user-2n} shows the accuracy at the top with noisy entries. HCoR performs the best. When one movie is recommended to each user, HCoR has an error rate of 18.4\% while AM has an error rate of 32.89\%. So the error rate of AM increases by 7.6\% comparing to the noise-free case while the error rate of HCoR increases only by 2.82\%.

\begin{figure}[htb]
\centering
\includegraphics[width=2.8in]{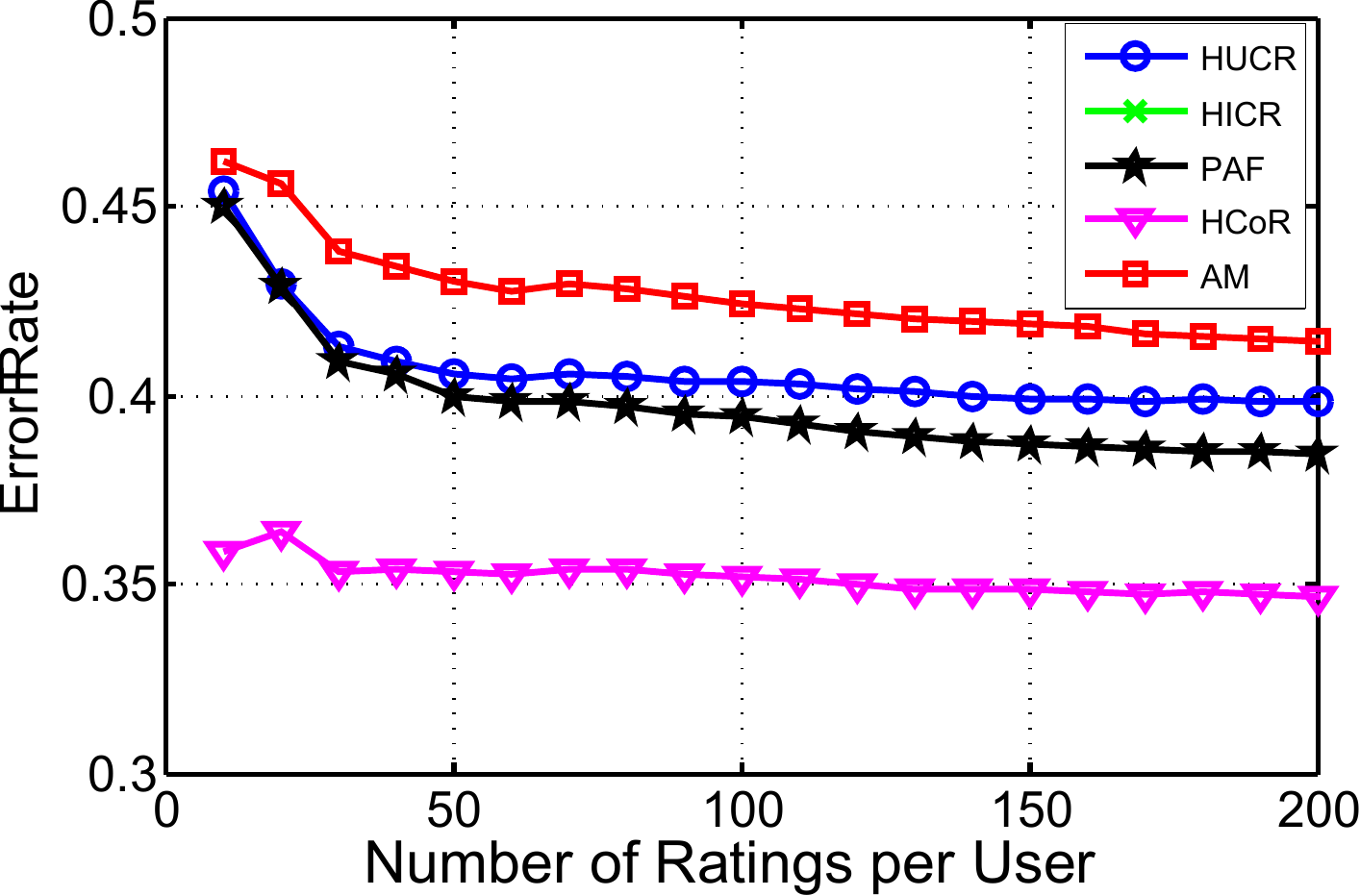}
\caption{Accuracy for information-sparse users for the Netflix data with noise. The figure shows the error rates of users who rate different numbers of movies.}
\label{fig: nf-30-70-sparse-u-2n}
\end{figure}

The results for the accuracy for information-sparse users is shown in Figure \ref{fig: nf-30-70-sparse-u-2n}. HCoR has the lowest error rate. For the users who rate less than 10 items, HCoR has an error rate of 35.88\% while the error rate of AM is 46.71\%. Comparing to the noise-free case, the error rate of HCoR increases by 1.93\% while the error rate of AM increases by 5.9\%.  HICR is not shown in the figure since the error rate is more than 50\%.

Table \ref{tab: er-nf-2n} summarizes the overall error rates. HCoR have the lowest overall error rate, which is 4.6\% lower than that of AM. Note that HCoR and AM have similar error rates in the noise-free case. From this set of experiments, we again see that HCoR is more robust to noise than AM.
\begin{table}[hbt]
\centering
\begin{tabular}{|c|c|c|c|c|}
  \hline
  HUCR  & HICR & HCoR & PAF & AM \\
\hline
 42.47\%& 43.51\%& 34.28\% & 39.06\% & 38.88\%\\
 \hline
\end{tabular}
\caption{The overall error rates for the Netflix dataset with noise} \label{tab: er-nf-2n}
\end{table}

\section{Proofs}
\label{sec: proofs}

\subsection{Notation}
\label{sec: notation}
\begin{itemize}[leftmargin=*]
\item $U:$ the number of users

\item $u$, $v,$ and $w:$ the user index and $u, v, w \in \{1, \cdots, U\}$

\item $M:$ the number of items

\item $m,$ $n,$ and $l:$  the item index and $m, n \in \{1, \cdots, M\}$

\item $K:$ the number of clusters

\item $k:$ the cluster index

\item $G:$ the number of preference (rating) levels

\item ${\bf B}:$ the preference matrix

\item ${\bf R}:$ the observed rating matrix

\item $\sigma_{u, v}:$ the similarity between user $u$ and user $v$

\item $\varphi_{u,v}:$ the number of items co-rated by users $u$ and $v$

\item $\sigma_{m,n}:$ the similarity between item $m$ and item $n$

\item $\varphi_{m,n}:$ the number of users who rate both items $m$ and $n$

\item $1-\alpha:$ the erasure probability of an information-sparse user (item)

\item $1-\beta:$ the erasure probability of an information-rich user (item)

\end{itemize}

Given non-negative functions $f(M)$ and $g(M),$ we also use the following order notation throughout the paper.
\begin{itemize}[leftmargin=*]
\item $f(M)=O(g(M))$ means there exist positive constants $c$ and $\tilde{M}$ such that $f(M) \leq cg(M)$ for all
$ M\geq \tilde{M}$.

\item $f(M)=\Omega(g(M))$ means there exist positive constants $c$ and $\tilde{M}$ such that $f(M)\geq
cg(M)$ for all $M\geq \tilde{M}.$ Namely, $g(M)=O(f(M))$.

\item $f(M)=\Theta(g(M))$ means  that both $f(M)=\Omega(g(M))$ and $f(M)=O(g(M))$ hold.

\item $f(M)=o(g(M))$ means that $\lim_{M\rightarrow \infty} f(M)/g(M)=0.$

\item $f(M)=\omega(g(M))$ means that $\lim_{M\rightarrow \infty} g(M)/f(M)=0.$ Namely,
$g(M)=o(f(M)).$
\end{itemize}

\subsection{Proof of Theorem \ref{thm: l-clu}}
\label{sec: proof-thm1}
Recall that an information-rich user's rating is erased with probability $1-\beta,$ an information-sparse user's rating is erased with probability $1-\alpha,$ and the number of information-rich users in each cluster is upper bounded by a constant $\eta.$

Since $\lim_{U\rightarrow \infty} \left(1-\frac{K}{U}\right)^{\frac{U}{K}} =e^{-1}$ and $U/K=\Omega(\log U),$  there exists a sufficiently large $\bar{U}$ such that for any $U\geq\bar{U},$
\begin{equation}\left(1-\frac{K}{U}\right)^{\frac{U}{K}}\geq e^{-1.1}.\label{eq: thm-1-prob}\end{equation}

Now consider the case $U\geq \bar{U}.$ If the theorem does not hold, then there exists a policy $\hat{\Phi}$ such that
\begin{equation}
\Pr(\hat{\Phi}({\bf R})={\bf B}|{\bf B}) > 1-\frac{\delta}{3}\label{eq: thm-1-hypo}
\end{equation} for all $\bf B$'s.

We define a set ${\cal R}$ such that ${\bf R}\in {\cal R}$ if in $\bf R,$ all ratings of the first item given by the users of the first cluster are erased. Note that
$$\Pr({\bf R}\in{\cal R}|{\bf B})\geq (1-\beta)^\eta (1-\alpha)^{\frac{U}{K}-\eta}=\left(\frac{1-\beta}{1-\alpha}\right)^\eta (1-\alpha)^{\frac{U}{K}}.$$ Given $\alpha\leq \frac{K}{U}$ and $U\geq \bar{U},$ we have
$$\Pr({\bf R}\in{\cal R}|{\bf B})\geq (1-\beta_{\max})^\eta e^{-1.1} =\delta,$$ and
\begin{equation}\Pr({\bf R}\not\in{\cal R}|{\bf B}) \leq 1-\delta.\label{eq: thm-1-p1}\end{equation}

Now given a preference matrix $\bf B,$ we construct $\hat{\bf B}$ such that it agrees with $\bf B$ on all entries except on the ratings to the first item given by the users in the first cluster. In other words, $\hat{b}_{nu}\not=b_{nu}$ if $n=1$ and $\lceil uK/U\rceil=1;$ and $\hat{b}_{nu}=b_{nu}$ otherwise. It is easy to verify that $\hat{\bf B}$ satisfies the fractionally separable condition for users (Condition (\ref{cond: frac-user})) since it changes only one moving rating for the users in the first cluster. Furthermore, for any ${\bf R}\in {\cal R},$ we have \begin{equation}\Pr\left({\bf R}|{\bf B}\right)=\Pr\left({\bf R}|\hat{\bf B}\right).\label{eq: thm-1-B}\end{equation}

Now we consider the probability of recovering $\hat{\bf B}$ under $\hat{\Phi},$ and have
\begin{eqnarray*}
&&\Pr\left(\left.\hat{\Phi}(\bf R)=\hat{\bf B}\right|\hat{\bf B}\right)\\
&=&\Pr\left(\left.(\hat{\Phi}(\bf R)=\hat{\bf B})\cap ({\bf R}\in{\cal R})\right|\hat{\bf B}\right)\\
&+&\Pr\left(\left.(\hat{\Phi}(\bf R)=\hat{\bf B})\cap ({\bf R}\not\in{\cal R})\right|\hat{\bf B}\right)\\
&\leq&\Pr\left(\left.(\hat{\Phi}(\bf R)\not={\bf B})\cap ({\bf R}\in{\cal R})\right|\hat{\bf B}\right)+\Pr\left(\left. {\bf R}\not\in{\cal R}\right|\hat{\bf B}\right)\\
&=_{(a)}&\Pr\left(\left.(\hat{\Phi}(\bf R)\not={\bf B})\cap ({\bf R}\in{\cal R})\right|{\bf B}\right)+\Pr\left(\left. {\bf R}\not\in{\cal R}\right|\hat{\bf B}\right)\\
&\leq &\Pr\left(\left.\hat{\Phi}(\bf R)\not={\bf B})\right|{\bf B}\right)+\Pr\left(\left. {\bf R}\not\in{\cal R}\right|\hat{\bf B}\right)\\
&\leq_{(b)}&\frac{\delta}{3}+1-\delta\\
&=&1-\frac{2\delta}{3}.
\end{eqnarray*} where equality $(a)$ holds due to equation (\ref{eq: thm-1-B}), and inequality $(b)$ yields from inequalities (\ref{eq: thm-1-hypo}) and (\ref{eq: thm-1-p1}). The inequality above contradicts (\ref{eq: thm-1-hypo}), so the theorem holds.

\subsection{Proof of Theorem \ref{thm: fea-clu}}
\label{sec: proof-thm2}
We first calculate the expectation of the similarity $\sigma_{uv}$ in the following cases:
\begin{itemize}[leftmargin=*]
\item Case 1: $u$ and $v$ are two different information-rich users in the same cluster. In this case, we have
\begin{eqnarray*}
E[\sigma_{uv}]&=&2M\beta^2\left(p^2+(G-1)\left(\frac{1-p}{G-1}\right)^2\right)-M\beta^2\\
&=&2M\beta^2\left(p^2+\frac{(1-p)^2}{G-1}\right)-M\beta^2,
\end{eqnarray*} where
$\beta^2$ is the probability the two users' ratings to item $m$ are not erased, $p^2$ is the probability that the observed ratings of the two users are their true preference, and $(G-1)\left(\frac{1-p}{G-1}\right)^2$ is the probability that the observed ratings of the two users are the same but not their true preference. We define $$z_1=\left(p^2+\frac{(1-p)^2}{G-1}\right),$$ so $E[\sigma_{uv}]=M\beta^2 (2z_1-1)$ in this case.

\item Case 2: $u$ and $v$ are in the same cluster, $u$ is an information-rich user, and $v$ is an information-sparse user. In this case, we have
\begin{eqnarray*}
E[\sigma_{uv}]&=&2M\alpha \beta\left(p^2+(G-1)\left(\frac{1-p}{G-1}\right)^2\right)-M\alpha\beta\\
&=&2M\alpha \beta\left(p^2+\frac{(1-p)^2}{G-1}\right)-M\alpha\beta\\
&=&M\alpha\beta (2z_1-1),
\end{eqnarray*} where $\alpha\beta$ is the probability the two users' ratings to item $m$ are not erased.

\item Case 3: $u$ and $v$ are in different clusters, and both are information-rich users. In this case, under the biased rating condition (\ref{cond: biased}), we can obtain
\begin{eqnarray*}
E[\sigma_{uv}]&\leq & 2\mu M \beta^2 z_1+2(1-\mu) M\beta^2\times\\
&&\left(2p\frac{1-p}{G-1}+(G-2)\left(\frac{1-p}{G-1}\right)^2\right)-M\beta^2\\
&=&2\mu M\beta^2 z_1 +2(1-\mu) M \beta^2\times\\
&&\left(\frac{1-p^2}{G-1}-\left(\frac{1-p}{G-1}\right)^2\right)-M\beta^2.
\end{eqnarray*} We define $$z_2=\left(\frac{1-p^2}{G-1}-\left(\frac{1-p}{G-1}\right)^2\right),$$ so $$M\beta^2 (2z_2-1)\leq E[\sigma_{uv}]\leq M\beta^2 (2\mu z_1+2(1-\mu) z_2-1)$$ in this case.

\item Case 4: $u$ and $v$ are in different clusters, $u$ is an information-rich user, and $v$ is an information-sparse user. In this case, we have
$$  M\alpha \beta (2z_2-1)\leq E[\sigma_{uv}]\leq M\alpha\beta (2 \mu z_1+2(1-\mu) z_2-1).$$

\item Case 5: $u$ and $v$ are in the same cluster, and are both information-sparse users. In this case, we have
$$  E[\sigma_{uv}]=M\alpha^2 (2z_1-1).$$

\item Case 6: $u$ and $v$ are in different clusters, and are both information-sparse users. In this case, we have
$$ E[\sigma_{uv}]\leq  M\alpha^2 (2\mu z_1+2(1-\mu)z_2-1).$$
\end{itemize}
Note that $z_1-z_2=\left(p-\frac{1-p}{G-1}\right)^2,$ so $z_1>z_2$ when $p> \frac{1}{G}.$ Now we define ${\cal P}_j$ to be the set of $(u,v)$ pairs considered in case $j$ above.

Recall that we assume $\alpha\beta M=\omega(\log U)$ and $\frac{\alpha}{\beta}=o(1).$ Given any $\epsilon>0,$ we define event ${\cal E}_j$ for $j\in\{1, 2, 3, 4\}$ to be $${\cal E}_j=\left\{(1-\epsilon)E[\sigma_{uv}]\leq \sigma_{uv}\leq (1+\epsilon)E[\sigma_{uv}] \quad \forall \ (u,v)\in {\cal P}_j\right\},$$ and ${\cal E}_j$ for $j=5,6$ to be
\begin{eqnarray*}
{\cal E}_5&=&\left\{\sigma_{uv}\leq 0.1 M\alpha\beta (2z_1-1)\quad \forall \ (u,v)\in {\cal P}_5\right\}\\
{\cal E}_6&=&\left\{\sigma_{uv}\leq 0.1 M\alpha\beta \left(2\mu z_1+2(1-\mu) z_2-1\right)\quad \forall \ (u,v)\in {\cal P}_6\right\},
\end{eqnarray*}

Using the Chernoff bound \cite[Thereorem 4.4 and Theorem 4.5]{MitUpf_05}, we now prove that when $M$ is sufficiently large,
\begin{equation}\Pr\left({\cal E}_j\right)\geq 1-\frac{1}{M}\label{eq: cher}\end{equation} for any $j.$ We establish this result by considering the following cases:
\begin{itemize}[leftmargin=*]
\item First consider Case 2 in which users $u$ and $v$ are in the same cluster. In this case, ${\bf 1}_{r_{vm}\not=\star, r_{um}\not=\star}$'s are identically and independently distributed (i.i.d.) Bernoulli random variables (across $m$). Applying the Chernoff bound and the fact that $|{\cal P}_j|\leq U^2$ for any $j,$ we  have
\begin{eqnarray*}
\Pr\left({\cal E}_2\right)
&=&\Pr\left(\left|{\sigma_{uv}-E[\sigma_{uv}]}\right|\leq \epsilon {E[\sigma_{uv}]},\ \forall \ (u,v)\in {\cal P}_2\right)\\
&\geq& 1-2U^2\exp\left(-\frac{\epsilon^2 E[\sigma_{uv}]}{3}\right)\\
&=& 1-2\exp\left(-\frac{\epsilon^2 M\alpha\beta(2z_1-1)}{3}+2\log U\right).
\end{eqnarray*} Since $\alpha\beta M=\omega (\log M)$ and $U=\Theta(M),$ when $M$ is sufficiently large, we obtain
\begin{eqnarray*}
\Pr\left({\cal E}_2\right)\geq 1-\frac{1}{M}.
\end{eqnarray*}

\item Next consider Cases 1, 3, 4, where the two users are in different clusters. Use Case 4 as an example. We assume users $u$ and $v$ have the same preference on items $1, \cdots, \mu_1M,$ and different preference on items $\mu_1M+1, \cdots, M,$ where $\mu_1<\mu.$ Then ${\bf 1}_{r_{vm}\not=\star, r_{um}\not=\star}$'s are i.i.d. Bernoulli random variables for $m=1, \cdots, \mu_1 M;$ and ${\bf 1}_{r_{vm}\not=\star, r_{um}\not=\star}$'s are i.i.d. Bernoulli random variables for $m=\mu_1M+1, \cdots, M.$ We can then prove inequality (\ref{eq: cher}) by applying the Chernoff bound to the two cases separately.

\item For Case 5, we define a new user $w$ who is in the same cluster with user $v$ and associated with an erasure probability $1-0.05\beta.$ Since $\alpha=o(\beta),$ we have for any $A>0,$ $$\Pr\left(\sigma_{wv} \geq A\right)\geq \Pr\left(\sigma_{uv}\geq A\right).$$ Then $\Pr\left({\cal E}_5\right)\geq 1-\frac{1}{M}$ can be proved by using the Chernoff bound to lower bound the probability that $\sigma_{wv}\leq 0.1 M\alpha\beta(2z_1-1).$ The proof for Case 6 is similar.
\end{itemize}

We further consider co-rating of two users $u$ and $v$ ($\varphi_{u,v}$) in the following two scenarios:
\begin{itemize}[leftmargin=*]
\item Scenario 1: $u$ and $v$ are both information-rich users. In this scenario, we have
\begin{eqnarray}
E[\varphi_{uv}]=M \beta^2.\label{eqn:sce1}
\end{eqnarray}

\item Scenario 2: $u$ is an information-rich user and $v$ is an information-sparse user. In this scenario, we have
\begin{eqnarray}
E[\varphi_{uv}]=M\alpha \beta.\label{eqn:sce2}
\end{eqnarray}
\end{itemize}
We now define ${\cal Q}_1$ to be the set of $(u,v)$ pairs in scenario 1, and ${\cal Q}_2$ to be the set of $(u,v)$ pairs in scenario 2.  We define  $${\cal F}_j=\left\{(1-\epsilon)E[\varphi_{uv}]\leq \varphi_{uv}\leq (1+\epsilon)E[\varphi_{uv}] \quad \forall \ (u,v)\in {\cal Q}_j\right\}$$ for $j=1,2.$
Based on the Chernoff bound, we have that when $M$ is sufficiently large for any $j,$
$$\Pr\left({\cal F}_j\right)\geq 1-\frac{1}{M}.$$

Without the loss of generality, we assume $2\mu z_1+2(1-\mu)z_2>1.$\footnote{The other cases can be proved following similar steps.} We choose $\epsilon\in(0, 1)$ such that
$$\frac{(1-\epsilon)^2}{(1+\epsilon)^2}\frac{2z_1-1}{2\mu z_1+2(1-\mu)z_2-1}>1.$$ Such an $\epsilon$ exists because $z_1>z_2.$
We further assume ${\cal E}_j$ $(j=1, 2, 3, 4, 5, 6)$ and ${\cal F}_j$ $(j=1,2)$ all occur. Now consider step (i) of the algorithm, if $u$ is an information-rich user, then the similarity between $u$ and $v$ is
\begin{eqnarray*}
\sigma_{uv}\left\{
              \begin{array}{ll}
                \geq (1-\epsilon) M\beta^2 (2z_1-1), & \hbox{ case 1;} \\
                \leq (1+\epsilon) M\beta^2 (2\mu z_1+2(1-\mu)z_2-1), & \hbox{ case 3;} \\
                \leq (1+\epsilon)M\alpha\beta (2z_1-1), & \hbox{ case 2;} \\
                \leq (1+\epsilon) M\alpha\beta (2\mu z_1+2(1-\mu)z_2-1), & \hbox{ case 4.}               \end{array}
            \right.
\end{eqnarray*}
Since $\sigma_{uv}$ is the largest when $v$ is an information-rich user in the same cluster, an information-rich user in the same cluster is picked in step (i) of the algorithm.

If $u$ is an information-sparse user, we have
\begin{eqnarray*}
\sigma_{uv}\left\{
              \begin{array}{ll}
                \geq (1-\epsilon) M\alpha \beta (2z_1-1), & \hbox{ case 2;} \\
                \leq (1+\epsilon) M\alpha \beta (2\mu z_1+2(1-\mu)z_2-1), & \hbox{ case 3;} \\
                \leq 0.1M\alpha\beta (2z_2-1), & \hbox{ case 5;} \\
                \leq 0.1 M\alpha\beta (2\mu z_1+2(1-\mu)z_2-1), & \hbox{ case 6.}               \end{array}
            \right.
\end{eqnarray*}
Again $\sigma_{uv}$ is the largest when $v$ is an information-rich user in the same cluster, so an information-rich user in the same cluster is picked in step (i) of the algorithm.

Now given $v$ is an information-rich user, based on equations (\ref{eqn:sce1}) and (\ref{eqn:sce2}), the normalized similarity $\tilde{\sigma}_{vw}$ satisfies
\begin{eqnarray*}
\tilde{\sigma}_{vw}\left\{
              \begin{array}{ll}
                \geq (1-\epsilon)^2 (2z_1-1), & \hbox{ case 1;} \\
                \geq (1-\epsilon)^2 (2z_1-1), & \hbox{ case 2;} \\
                \leq (1+\epsilon)^2 (2\mu z_1+2(1-\mu)z_2-1), & \hbox{ case 3;} \\
                \leq (1+\epsilon)^2 (2\mu z_1+2(1-\mu)z_2-1), & \hbox{ case 4.}               \end{array}
            \right.
\end{eqnarray*}
So the normalized similarity when $w$ is in the same cluster as $v$ is larger than the similarity when $w$ is not in the same cluster. Therefore in step (i) of the algorithm, all users in the same cluster as $v$ are selected. $v$ and $u$ are in the same cluster, so at the end of step (ii), all users are in user $u$'s cluster are selected.

Now consider the ratings of item $m$ given by user-cluster $k$ and define $$M_{m,k,g}=\sum_{u: \lceil uK/U\rceil=k} {\bf 1}_{r_{um}=g}$$ to be the number of users in cluster $k$ who give $g$ to item $m.$  With a slight abuse of notation, let $b_{km}$ to be the true preference of users in cluster $k$ to item $m,$ so we have
\begin{eqnarray*}
E\left[M_{m,k,g}\right]
\left\{
                                                             \begin{array}{ll}
                                                               \geq \left(2\beta+(\frac{U}{K}-2)\alpha\right) p, & g=b_{km}\\
                                                               \leq \left(\eta\beta+(\frac{U}{K}-\eta)\alpha\right) \frac{1-p}{G-1}, & g\not=b_{km}
                                                             \end{array}
                                                           \right.
\end{eqnarray*}

Define $\cal G$ to be the event that a majority voting within a user-cluster gives the true preference of an item for all items and user-clusters, i.e.,
$${\cal G}=\{b_{km}=\arg\max_g M_{m,k,g}\quad \forall \ m, k\}.$$
Now when $\frac{\alpha U}{K}=\omega(\log M),$ using the Chernoff bound, it is easy to verify that
$$\Pr\left({\cal G}\right)\geq 1-\frac{1}{M}.$$

Now when ${\cal E}_j$ $(j=1, 2, 3, 4, 5)$ and ${\cal F}_j$ $(j=1, 2)$ occur, the users are clustered correctly by the algorithm; and when ${\cal G}$ occurs, a majority voting within the cluster produces the true preference. Therefore, the theorem holds.

\subsection{Proof of Theorem \ref{thm: l-co-clu}}
\label{sec: proof-thm3}

Given $\alpha\leq \frac{K^2}{UM},$ $\beta \leq \frac{K}{\eta(M+U)-\eta^2 K}$ and a constant $\eta,$ there exists $\bar{M}$ such that for any $M\geq \bar{M},$ $$(1-\alpha)^{\left(\frac{U}{K}-\eta\right)\left(\frac{M}{K}-\eta\right)}\geq  e^{-1.1},$$ and $$(1-\beta)^{\frac{\eta M}{K}+\frac{\eta U}{K}-\eta^2}\geq e^{-1.1}.$$

Now consider the case $M\geq \bar{M}$ and $K=\Theta(\log M).$ If the theorem does not hold, then there exists a policy $\hat{\Phi}$ such that
\begin{equation}
\Pr(\hat{\Phi}({\bf R})={\bf B}|{\bf B})> 1-\frac{\delta}{3}\label{eq: thm-2-hypo}
\end{equation} for all $\bf B$'s.

We define a set ${\cal R}$ such that ${\bf R}\in {\cal R}$ if in $\bf R,$ all ratings of the items in the first item-cluster given by the users of the first cluster are erased. Note that when $M\geq \bar{M},$ we have
\begin{eqnarray}
\Pr({\bf R}\in {\cal R}|{\bf B})\nonumber&\geq &(1-\beta)^{\frac{\eta M}{K}+\frac{\eta U}{K}-\eta^2} (1-\alpha)^{\left(\frac{U}{K}-\eta\right)\left(\frac{M}{K}-\eta\right)}\nonumber\\
&\geq& e^{-2.2}=\delta.\label{eq: thm-2-p}
\end{eqnarray}

Now given a preference matrix $\bf B,$ we construct $\hat{\bf B}$ such that it agrees with $\bf B$ on all entries except on the rating to the first item-cluster given by the first user-cluster. In other words, $\hat{b}_{nu}\not=b_{nu}$ if $\lceil nK/M\rceil=1$ and $\lceil uK/U\rceil=1;$ and $\hat{b}_{nu}=b_{nu}$ otherwise. It is easy to verify that $\hat{\bf B}$ satisfies the fractionally separable conditions both for users and items (Conditions (\ref{cond: frac-user}) and (\ref{cond: frac-item})) as long as $\hat{\bf B}$ satisfies the two fractionally separable conditions because the construction of $\hat{\bf B}$ changes only the rating of the first (item-cluster, user-cluster) pair and $K=\Theta(\log M).$ Furthermore, for any ${\bf R}\in {\cal R},$ we have \begin{equation}\Pr\left({\bf R}|{\bf B}\right)=\Pr\left({\bf R}|\hat{\bf B}\right).\label{eq: thm-1}\end{equation} Following the same argument in the proof of Theorem \ref{thm: l-clu}, we have
\begin{eqnarray*}
\Pr\left(\left.\hat{\Phi}(\bf R)=\hat{\bf B}\right|\hat{\bf B}\right)&\leq &\Pr\left(\left.\hat{\Phi}(\bf R)\not={\bf B})\right|{\bf B}\right)+\Pr\left(\left. {\bf R}\not\in{\cal R}\right|\hat{\bf B}\right)\\
&\leq&\frac{\delta}{3}+1-\delta\\
&=&1-\frac{2\delta}{3},
\end{eqnarray*} which contradicts (\ref{eq: thm-2-hypo}). So the theorem holds.

Note that $E[X_{\bf R}]\geq \alpha UM$ and $E[X_{\bf R}]\geq \beta (\eta K (M+U)- \eta^2K^2)$ always hold. So $E[X_{\bf R}] \leq K^2$ implies $\alpha\leq \frac{K^2}{UM}$ and $\beta \leq \frac{K}{\eta(M+U)-\eta^2 K}.$

\subsection{Proof of Theorem \ref{thm: fea-co-clu}}
\label{sec: proof-thm4}
Following similar argument as the proof of Theorem \ref{thm: fea-clu}, we can prove that when $\alpha\beta M=\omega(\log M)$ or $\beta^2 K=\omega(\log M)$ all user-clusters and item-clusters are correctly identified with probability at least $1-\frac{1}{M}.$

Now consider the ratings of item-cluster $k_m$ given by user-cluster $k_u$ and define $$M_{k_u,k_m,g}=\sum_{u: \lceil uK/U\rceil=k_u}\sum_{m: \lceil mK/M\rceil=k_m} {\bf 1}_{r_{um}=g}$$ to be the number of $g$ ratings given by users in cluster $k_u$ to to items in cluster $k_m.$  With a slight abuse of notation, let $b_{k_u,k_m}$ to be the true preference. Further let $\eta_{k_u}$ denote the number of information-rich users in cluster $k_u$ and  $\eta_{k_m}$ denote the information-rich items in cluster $k_m.$ When $g=b_{k_u,k_m},$ we have
\begin{eqnarray*}
E\left[M_{k_u,k_m,g}\right]&=&\left(\left(\eta_{k_u}\frac{M}{K}+\eta_{k_m}\frac{U}{K}-\eta_{k_u}\eta_{k_m}\right)\beta+\right.\\
&&\left.\left(\frac{U}{K}-\eta_{k_u}\right)\left(\frac{M}{K}-\eta_{k_m}\right)\alpha\right)  p;
\end{eqnarray*}
and otherwise,
\begin{eqnarray*}
E\left[M_{k_u,k_m,g}\right]&=&\left(\left(\eta_{k_u}\frac{M}{K}+\eta_{k_m}\frac{U}{K}-\eta_{k_u}\eta_{k_m}\right)\beta+\right.\\
&&\left.\left(\frac{U}{K}-\eta_{k_u}\right)\left(\frac{M}{K}-\eta_{k_m}\right)\alpha\right) \frac{1-p}{G-1}.\end{eqnarray*}

Define $\cal G$ to be the event that a majority voting within an item and user-cluster gives the true preference of item-cluster for all items and user-clusters, i.e.,
$${\cal G}=\{b_{k_u, k_m}=\arg\max_g M_{k_u,k_m,g}\quad \forall \ k_u, k_m\}.$$
Now when $\frac{\alpha UM}{K^2}=\omega(\log M)$ or $\frac{\beta M}{K}=\omega(\log M),$ using the Chernoff bound, it is easy to verify that
$$\Pr\left({\cal G}\right)\geq 1-\frac{1}{M}.$$

Further, $E[X_{\bf R}]=\omega(K^2 \log M)$ implies that $\frac{\alpha UM}{K^2}=\omega(\log M)$ or $\frac{\beta M}{K}=\omega(\log M),$ so the theorem holds.

\section{Conclusion}
In this paper, we considered both the clustering and co-clustering models for collaborative filtering in the presence of information-rich users and items. We developed similarity based algorithms to exploit the presence of information-rich entities. When users/items are clustered, our clustering algorithm can recover the rating matrix with $\omega(MK\log M)$ noisy entries; and when both users and items are clustered, our co-clustering can recover the rating matrix with $K^2\log M$ noisy entries when $K$ is sufficiently large. We compared our co-clustering algorithm with PAF and AM by applying them to the MovieLens and Netflix data sets. In the experiments, our proposed algorithm HCoR has significantly lower error rates when recommending a few items to each user and when recommending items to the majority of users who only rated a few items. Due to space limitations, we only presented the proofs for the basic models in Section \ref{sec: proofs}. The extensions mentioned in the remarks in Section \ref{sec: model} are straightforward. Furthermore, instead of assuming the cluster size is given in the clustering and co-clustering algorithms, the algorithms can estimate the erasure probability $\alpha$ using the number of observed ratings, i.e., $\alpha=1-\frac{{\bf X}_{\bf R}}{UM},$ from which, the algorithms can further estimate the cluster-size. This estimation can be proved to be asymptotically correct.\vspace*{-3pt}

\bibliographystyle{IEEEtran}
\bibliography{reference_all}

\end{document}